%% file: 0__main.tex
\documentclass[final,3p,times,twocolumn]{elsarticle} 

\usepackage{amssymb} 
\usepackage{amsmath} 
\usepackage{amsthm} 

\theoremstyle{definition}
\newtheorem{definition}{Definition}[section]

\usepackage{microtype}
\usepackage{lmodern}

\usepackage{subfig} 
\usepackage{enumitem} 

\usepackage{booktabs}
\usepackage{multirow}

\usepackage{adjustbox}

\usepackage{algorithm} 
\usepackage{algpseudocode} 
\floatname{algorithm}{Algorithm}
\algrenewcommand\algorithmicrequire{\textbf{Input:}}
\algrenewcommand\algorithmicensure{\textbf{Output:}}

\DeclareMathOperator*{\argmax}{arg\,max}
\DeclareMathOperator*{\argmin}{arg\,min}

\usepackage{svg}

\usepackage{bm}

\usepackage{hyperref} 

\begin{document}

\begin{frontmatter}



\title{Federated Clustering: An Unsupervised Cluster-Wise Training for Decentralized Data Distributions}

\author[sns,iit]{Mirko Nardi}
\ead{mirko.nardi@sns.it}

\author[iit]{Lorenzo Valerio}
\ead{lorenzo.valerio@iit.cnr.it}

\author[iit]{Andrea Passarella}
\ead{andrea.passarella@iit.cnr.it}

\affiliation[sns]{organization={Scuola Normale Superiore},
            city={Pisa},
            country={Italy}}
            
\affiliation[iit]{organization={IIT-CNR},
            city={Pisa},
            country={Italy}}

\input{00_abstract}




\end{frontmatter}



\input{01_introduction}
\input{02_related_works}

\input{03_problem_and_assumption}
\input{04_method}
\input{05_experiments}

\input{05_results}

\input{06_conclusion}

\section*{Acknowledgement}
This work was partially supported by the H2020 HumaneAI Net (952026) and CNR-STRIVE-MeSAS. A. Passarella's work was partly funded by the PNRR - M4C2 - Investimento 1.3, Partenariato Esteso PE00000001 - "RESTART", both funded by the European Commission under the NextGeneration EU programme.



\bibliographystyle{elsarticle-num-names}  
 \bibliography{bib_msn,bib_fl_clust_ref, bib_fl_rel}





%

\end{document}

%% file: 00_abstract.tex
\begin{abstract}
Federated Learning (FL) enables decentralized machine learning while preserving data privacy, making it ideal for sensitive applications where data cannot be shared. While FL has been widely studied in supervised contexts, its application to unsupervised learning remains underdeveloped. This work introduces FedCRef, a novel unsupervised federated learning method designed to uncover all underlying data distributions across decentralized clients without requiring labels. This task, known as Federated Clustering, presents challenges due to heterogeneous, non-uniform data distributions and the lack of centralized coordination.
Unlike previous methods that assume a one-cluster-per-client setup or require prior knowledge of the number of clusters, FedCRef generalizes to multi-cluster-per-client scenarios. Clients iteratively refine their data partitions while discovering all distinct distributions in the system. The process combines local clustering, model exchange and evaluation via reconstruction error analysis, and collaborative refinement within federated groups of similar distributions to enhance clustering accuracy.
Extensive evaluations on four public datasets (EMNIST, KMNIST, Fashion-MNIST and KMNIST49) show that FedCRef successfully identifies true global data distributions, achieving an average local accuracy of up to 95\%. The method is also robust to noisy conditions, scalable, and lightweight, making it suitable for resource-constrained edge devices.

\end{abstract}

%% file: 01_introduction.tex
\section{Introduction}

Distributed machine learning at the edge is recognized as an effective strategy to address the limitations of centralized ML systems, including privacy risks, data sovereignty issues, and the high data transmission costs of collecting vast amounts of data centrally~\citep{Verbraeken2020}. Among various distributed approaches, Federated Learning (FL) stands out \citep{mcmahan_communication-efficient_2017, konevcny2016federated}, due to its effectiveness in training ML models (typically NN-based) preserving data locality. Specifically, FL protocols involve a set of nodes (or \textit{clients}) that individually train a model locally on their own data, and only the model updates are shared. Hence, a global model is iteratively aggregated and distributed to all nodes until convergence. Typically, this process is coordinated by a central server, but serverless protocols have also been proposed~\citep{lalitha2018fully,beltran2023decentralized,yuan2024decentralized}.

While FL was initially designed for communication efficiency, significant progress has been made in addressing its well-known challenges~\citep{kairouz2021advances}, primarily including privacy and security~\citep{wei2020federated,Mothukuri2021Feb}, adaptation to heterogeneous data distributions (e.g., non-IID)~\citep{zhao2018federated-noniid} and variability in device capabilities~\citep{khan2020federated}.
However, most FL research still focuses on supervised learning  \citep{Singh2022Feb}. 

In many real-world applications, labeled data is scarce or unavailable, and labeling is prohibitively expensive. This highlights the need for expanding FL to unsupervised learning settings \citep{kairouz2021advances}, raising significant challenges in the learning process due to the lack of explicit guidance from labels.
Recent advancements in Unsupervised Federated Learning (UFL) have explored two main directions~\citep{Ji2021Feb,Jin2023Mar}. First, the integration of Federated Learning (FL) with Unsupervised Representation Learning (URL), typically through self-supervised learning frameworks such as contrastive learning or predictive coding. These approaches aim to extract general-purpose feature representations from decentralized data without relying on labels. Although effective in enhancing downstream performance, they often require deep architectures, negative sampling strategies, and large batch sizes—resulting in high computational and memory demands that are not well-suited for resource-constrained edge devices commonly found in FL systems.
Second, the clustering of unlabeled data within a federated framework, known as Federated Clustering (FC), whose goal is to identify patterns and group samples with similar features without data centralization, thereby maintaining data privacy and locality.
Among the two approaches, Federated Clustering (FC) remains significantly less studied and our work specifically addresses this gap. Existing works rely on extending traditional clustering algorithms such as $K$-means~\citep{dennis2021heterogeneity-kfed} or make restrictive assumptions about data distributions (e.g., fixed/limited number of clusters to discover), limiting their applicability in real-world scenarios~\citep{gosh2020IFCA,dennis2021heterogeneity-kfed}.
In this regard, Section~\ref{sec:related} provides a detailed discussion of related work.

This work introduces a more generalized approach to Federated Clustering, designed to address real-world decentralized learning scenarios where the data distributions are unevenly spread across the system.  
We assume that individual clients might observe only a subset of the distributions present in the system and lack explicit knowledge about their identities or how they relate to the broader global structure. Moreover, the clients can only operate on local data. 
The final goal is to discover the true global categories without requiring direct data sharing. In this context, we define their total number as \textit{global} $K_G$.

A real-world example of unsupervised federated clustering arises in healthcare, where hospitals collect local datasets—such as medical images or diagnostic records—related to different subsets of conditions. Due to privacy constraints, they cannot share raw patient data. Each hospital performs local clustering to group similar cases without knowing the full spectrum of diseases across the network. Our approach enables them to identify and align shared clusters collaboratively—representing distinct disease types—without exchanging raw data. In this context, the global number of clusters $K_G$ corresponds to the complete set of underlying medical categories the system aims to uncover.

Similar challenges emerge in other decentralized sensing environments, such as IoT networks or industrial systems, where each device captures local behavioral or operational data—whether images, sensor traces, or other modalities. Devices may independently observe and cluster recurring patterns, but lack awareness of whether these patterns are globally common or unique.

Our approach, enables such devices to collaboratively uncover the complete set of underlying data distributions in a decentralized manner, without requiring prior knowledge of the number of distributions and without sharing raw data. 
Each client begins by clustering its local data and trains a representation model for each cluster. These models are then exchanged among clients, allowing them to assess similarity and identify which clusters correspond to the same underlying distribution. Groups of clients with matching clusters engage in federated training to build shared models for those distributions. These models are then used to refine local cluster splits, improving alignment across the network. This process is repeated iteratively, enabling clients to progressively uncover the global structure.

The key intuition behind our approach is that only \textit{well-formed} clusters---those consisting of samples from the same underlying data distribution---should form associations. Through iterative refinement, these clusters become increasingly \emph{homogeneous}, leading to more reliable associations.

Overall, our main contributions are:
\begin{itemize}
  \item A generalized framework for Unsupervised Federated Clustering that identifies all the distinct data distributions in the system.
  \item  A novel, iterative federated learning strategy operating \emph{cluster-wise}, i.e., where clients contribute to multiple federated models by partitioning their local data, unlike other approaches that train on the entire local dataset.
  \item Evaluation on real-world non-i.i.d. data distributions,  where clients possess datasets with a variable number of data distributions (local $K$). This assumption is not addressed in existing literature, to the best of our knowledge.
\end{itemize}

Our previous work~\citep{nardi2022flanomaly} demonstrated the effectiveness of FL in anomaly detection, where models collaboratively identified outliers (i.e., identifying two groups of data, ``normal" and ``anomalous" data) across decentralized datasets. Here, we generalize this approach to a broader unsupervised setting, enabling fully decentralized discovery of multiple data distributions which can be an arbitrary number, none of which anomalous (even though the proposed algorithm can be used for anomaly detection as a special case).

The remainder of this paper is organized as follows. Section~\ref{sec:related} provides a thorough and comprehensive review of related work in the field of federated learning, unsupervised learning, and their intersection in UFL, highlighting the key challenges and existing approaches.
Section~\ref{sec:problem_assumptions} introduces the problem statement and the assumptions we make in order to frame the task. Section~\ref{sec:methodology} details our proposed methodology, including the algorithmic framework and the rationale behind our two-fold approach. Section~\ref{sec:experiments} presents the experimental setup, datasets used, and performance evaluation metrics and Section~\ref{sec:results} discusses the results and insights gained from our experiments. Finally, Section~\ref{sec:conclusion} concludes the paper with a summary of our findings and directions for future research.

%% file: 02_related_works.tex
\section{Related work}
\label{sec:related}

Federated Learning has seen substantial development in supervised and personalized settings, where the goal is to train global or client-specific models under privacy constraints. However, clustering in fully unsupervised federated environments remains underexplored.

Early approaches focused on \textit{clustered federated learning}, which aims to group clients with similar data distributions~\citep{smith2017federated, ghosh2019robust, sattler2020clustered}. However, these methods primarily target personalization by training shared models within each client group to improve downstream supervised task performance.

A notable recent contribution to this direction is IFCA \citep{gosh2020IFCA}, which operates by alternating between assigning clients to clusters and collectively training specialized models for each group. Building on this work, the same authors proposed SR-FCA \citep{vardhan2024improved-srfca}, which addresses IFCA's limitations by  eliminating specific initialization requirements and by allowing the same users in a cluster to have non-identical models (or data distributions).

Both approaches organize clients based on data characteristics---typically introduced through image rotations---to form groups with similar distributions. Within each group, the goal is to train specialized models for supervised tasks; while effective for personalization, these methods fundamentally use clustering as a tool to improve downstream supervised performance rather than being an unsupervised objective.

An unsupervised federated clustering approach is introduced in K-Fed~\citep{dennis2021heterogeneity-kfed}, which extends classical $K$-means~\citep{lloyd1982least-kmeans} to the federated setting by collecting local cluster centers and applying centralized $K$-means over them. However, the method presents several limitations: it assumes knowledge of both the global number of clusters $K$ and each client’s local number of clusters; it is non-iterative and centralized; and it does not leverage deep representations, making it less suitable for complex, high-dimensional data (e.g., images). 
In fact, K-Fed is primarily tested on synthetic data and \textit{well-separated} distributions. When applied to MNIST, the evaluation is limited to client-level clustering for personalization of supervised FL tasks rather than clustering individual data points, which limits its applicability.
Nonetheless, we take inspiration from K-Fed’s observation that data heterogeneity across clients can be beneficial. In particular, they assume that each client observes a subset of the global clusters, which aligns with our assumption of partial local knowledge in decentralized systems.

A more recent contribution to unsupervised federated clustering is UIFCA \citep{chung2022federated-uifca}, which extends IFCA by allowing each client to hold data from multiple clusters. The method trains K generative models—one per cluster—and assigns each datapoint to the model yielding the highest likelihood. Unlike prior approaches that focus on clustering personalization for supervised tasks, UIFCA shifts the focus to fully unsupervised clustering using unlabeled data.
However, UIFCA still imposes key limitations: the number of global clusters K is assumed to be known, and the non-IID data distribution is simplified, i.e., each client holds a dominant cluster and random samples from others. Additionally, the number of clients is fixed to match the number of clusters, limiting scalability. While effective on rotated-digit clustering, UIFCA shows poor performance when the clustering task is based on full-digit categories in MNIST, highlighting limitations in generalization.

In summary, while the listed clustered federated learning methods address non-IID distributions by forming groups of clients, they remain primarily oriented toward supervised downstream tasks or rely on simplifying assumptions. Our work differs in focusing on fully unsupervised federated clustering without requiring labels or restrictive initialization strategies, thereby extending applicability to more realistic and heterogeneous scenarios.




%% file: 03_problem_and_assumption.tex
\section{Problem Statement and System Assumptions}
\label{sec:problem_assumptions}

\begin{figure*}[ht] 
    \centering
    \includegraphics[width=0.65\textwidth]{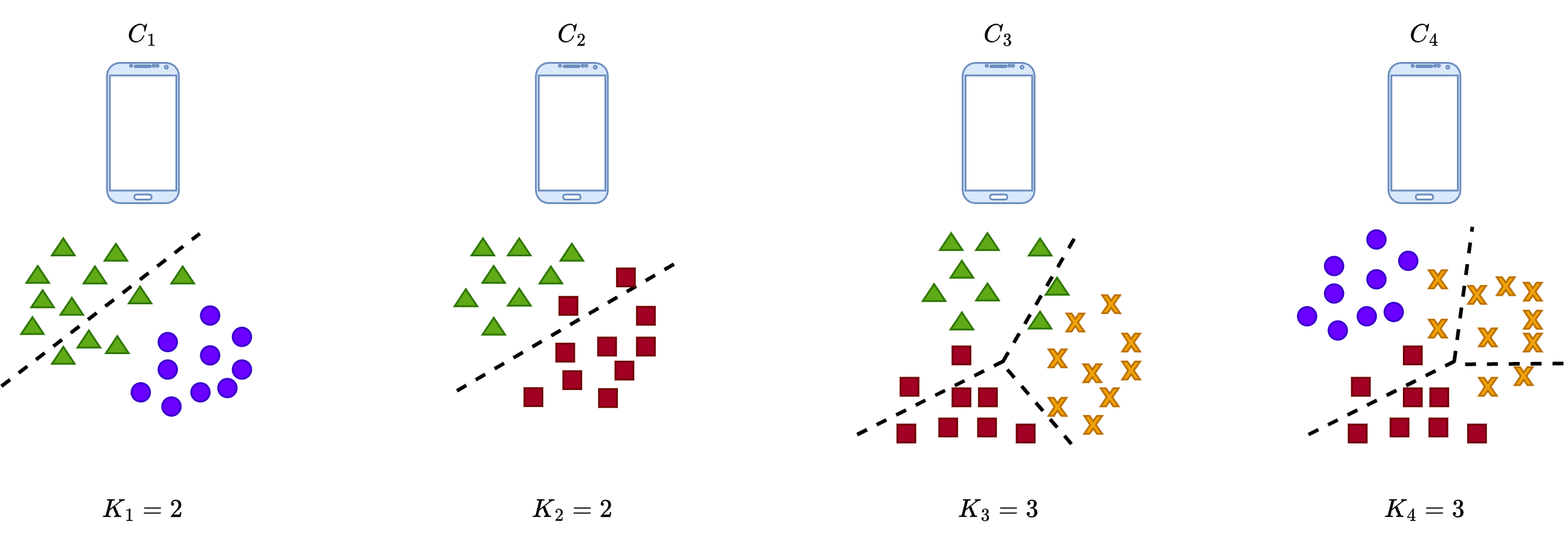}
    \caption{Schematic representation of the considered federated learning system with $N=4$ clients. 
    The objective is to identify the set of unique global data distributions $U$ (global $K_G=4$) represented by distinct shapes.
    Within each client $C_i$, $K_i$ denotes the number of unique data distributions present in the local dataset.
    The dotted lines indicate a possible clustering split $Q_i$, highlighting how imperfect local clusters may be formed within each client’s dataset.}
   \label{fig:system_architecture}
\end{figure*}


We consider a federated system consisting of $N$ clients, where each client $C_i$ holds a local dataset $D_i$. The global dataset $D$ is the union of all local datasets:
\begin{equation}
D = \bigcup_{i=1}^{N} D_i.
\end{equation}

In the system, there are totally $K_G$ underlying data distributions (or true categories) we aim to find, denoted as the set $U = \{ U_1, U_2, \dots, U_{K_G} \}$. 
The $i$-th client holds only a subset of the true global categories in its local dataset, denoted as $U_i \subset U$, where the number of categories in $U_i$ is given by $K_i$, with $2 \leq K_i < K_G$. 
This constraint mirrors real-world situations where data is naturally partitioned and incomplete. In supervised contexts, this phenomenon is referred to as \textit{label distribution skew},  a well-known type of non-IID data distribution~\cite{kairouz2021advances}.
Additionally, clients may share some of the same underlying distributions. The described system is illustrated in Fig.~\ref{fig:system_architecture}.

Since we assume an unsupervised setting, clients do not know their actual categories. Instead, they approximate them by performing unsupervised clustering on their local datasets, grouping data points based on intrinsic structure. Importantly, our methodology is orthogonal to the specific clustering algorithm used at the client side—any unsupervised method can be applied to initialize local clusters, as our approach operates independently of this choice. The set of clusters identified by the $i$-th client is denoted as $Q_i = \{ Q_{i1}, Q_{i2}, \dots, Q_{iK_i} \}$, where each $Q_{ij}$ represents a local cluster.

We recall that in clustering literature, the parameter $K$ typically denotes the number of partitions into which data points are grouped. In our federated system, we distinguish between: \textit{local} $K_i$, the number of clusters identified within the dataset of client $C_i$; \textit{global} $K_G$, the total number of unique data distributions across all clients, which must be inferred collaboratively. Both local and global $K$ are to be estimated in this system, and they are not known apriori. 
Additionally, we have the following assumptions:
(i) our system operates under a decentralized, \textit{serverless} federated learning protocol, where clients communicate by exchanging only model parameters to enforce data privacy. 
(ii) Each client is assumed to have the computational capability to train local models but within the constraints of edge computing environments, which have more limited resources compared to centralized systems. 
(iii) Furthermore, we strictly follow an unsupervised learning paradigm, meaning that no apriori assigned labels are used at any stage of the process. 

Given this setting, our proposed methodology is designed to achieve three key objectives.
\paragraph{Objective 1:} FedCRef aims to identify all global data distributions  $U = \{ U_1, U_2, \dots, U_{K_G} \}$ across the network without prior knowledge of their total number ($K_G$) and direct data sharing. Alongside this discovery, the method constructs a set of compressed representation models $M = \{ M_1, M_2, \dots, M_{K_G} \}$, one for each identified distribution, allowing efficient knowledge transfer among clients.
\paragraph{Objective 2:} FedCRef seeks to improve the accuracy of local clustering at each client. Specifically, it aims to align the local cluster assignments $Q_i$ with the true underlying categories $U_i$. This alignment is assessed through the progressive improvement of clustering quality over iterations, measured using unsupervised clustering accuracy (ACC). Note that true labels (e.g., the shapes in Fig.~\ref{fig:system_architecture}) are used solely for evaluation and are never accessed during training.
\paragraph{Objective 3:} FedCRef explicitly addresses non-IID data distributions by supporting scenarios in which clients hold only a subset of the global distributions, i.e., $2 \leq K_i < K_G$. 

%% file: 04_method.tex
\section{Methodology}
\label{sec:methodology}

\begin{algorithm}
\footnotesize
\caption{Federated Cluster Refinement}
\label{alg:fed_complete}
\begin{algorithmic}[1]
\Procedure {FedCRef}{Clients set $C$, percentile $\alpha$, association threshold $\theta$, stability threshold $\tau$}
\State Let $D_i$ be the local dataset of client $C_i$ \label{alg1:line:step:1:start}
\State Let $Q_i = \{Q_{i1},\dots, Q_{iK_i}\}$ be the clusters of $C_i$ \label{alg1:line:init_clusters}
\State Let $M_i = \{M_{i1},\dots, M_{iK_i}\}$ be the models of $C_i$ clusters 
\State Initialize $A = C$ as the set of active Clients
\While{(varying $I$ and $G$) or $|A| > 0$}  \label{alg1:line:convergence}
    \For{each $C_i \in A $ and each $Q_{iq} \in Q_i$} 
        \State Train model $M_{iq}$ on cluster $Q_{iq}$ \label{alg1:line:local_train}
    \EndFor \label{alg1:line:step:1:end}
    \State All-to-all exchange $M_{iq}$ models among $C_i \in C$ \label{alg1:line:step:2:start}
    \For{each pair of clusters $(Q_{iq}, Q_{jp})$ where $Q_{iq}$ $\in$ $C_i$ and $Q_{jp}$ $\in$ $C_j$}
        \State $D1_{\text{norm}}= \text{Normalize}_{[0,1]}|E_{iq \rightarrow iq} - E_{jp \rightarrow iq}|$
        \State $D2_{\text{norm}} = \text{Normalize}_{[0,1]}|E_{jp \rightarrow jp} - E_{iq \rightarrow jp}|$
        \If{
        $Q_{\alpha}(D1_{\text{norm}}) \leq \theta\ \geq \alpha$ and
        $Q_{\alpha}(D2_{\text{norm}}) \leq \theta\ \geq \alpha$
        } \label{line:condition_link}
        \State Associate $Q_{iq}$ and $Q_{jp}$ 
        \EndIf
    \EndFor \label{alg1:line:step:2:end}
    \State Share associations links and initialize $\Gamma$ as the undirected graph of clusters  \label{alg1:line:step:3:start} 
    \State Let $\mathcal{G}$ be the connected component (c.c.)
    \State Let $G = \{G_i \in \mathcal{G} : |V_{G_i}| \geq 2, i= 1,2,\dots, |\mathcal{G}|\} $ set of c.c. with at least two cluster
    \State Let $I$ be the set of Isolated clusters (isolated nodes)
    \For{each connected component $G_p$ in $G$}
        \State Train model $M_{G_p}$ using FL on $G_p$ clusters
    \EndFor \label{alg1:line:step:3:end} 
    \State Let $M_G$ the set of all trained $M_{G_p}$ \label{alg1:line:step:4:start} 
    \State Distribute $M_{G}$ to all $C_i$
    \For{each $C_i \in A$} \label{alg1:line:cluster_refine}
        \State ClusterRefine($C_i$, $M_G$)
        \If{$ \text{ACC}( y'_{C_i}, y_{C_i}) \geq \tau $} 
        \State $A = A \setminus C_i$
        \EndIf
    \EndFor \label{alg1:line:step:4:end}
\EndWhile
\EndProcedure
\end{algorithmic}
\end{algorithm}

This section presents our methodology for addressing the identified objectives, as outlined in Algorithm~\ref{alg:fed_complete}.
Our proposed algorithm unfolds in four sequential and iterative phases: (1) each client independently performs local clustering and trains a representation model for each identified cluster, summarizing its structure without labels. (2) Clients exchange their trained models and assess pairwise similarities through reconstruction error, identifying which clusters across clients are likely to originate from the same underlying data distribution. This leads to the phase (3) where clients form temporary federated groups based on these associations and collaboratively train refined models for the shared distributions. (4) Finally, clients use these improved models to refine their local cluster assignments. This process is repeated iteratively, allowing clusters to progressively align across clients and converge toward the true global distributions.

\subsection*{Phase 1: Local Clusters Model Training} 
This step covers lines~\ref{alg1:line:step:1:start}-\ref{alg1:line:step:1:end} of Algorithm~\ref{alg:fed_complete}.
Considering that each client $C_i$ holds a local dataset $D_i$ partitioned into $K_i$ clusters, we assume an initial clustering estimation as the starting condition---this can be obtained using any suitable unsupervised algorithm tailored to the data characteristics (line~\ref{alg1:line:init_clusters}), as this step is orthogonal to our methodology.
Here, each client independently trains models to learn compressed representations of its local clusters. This process serves two main objectives: 
(i)~Clients are enabled to share information about their clusters without exposing raw data;
(ii) The learned representations are exploited to establish relationships between clients' data by identifying similarities in their local data distributions. 

\begin{figure}
    \centering
    \includegraphics[width=0.35\linewidth]{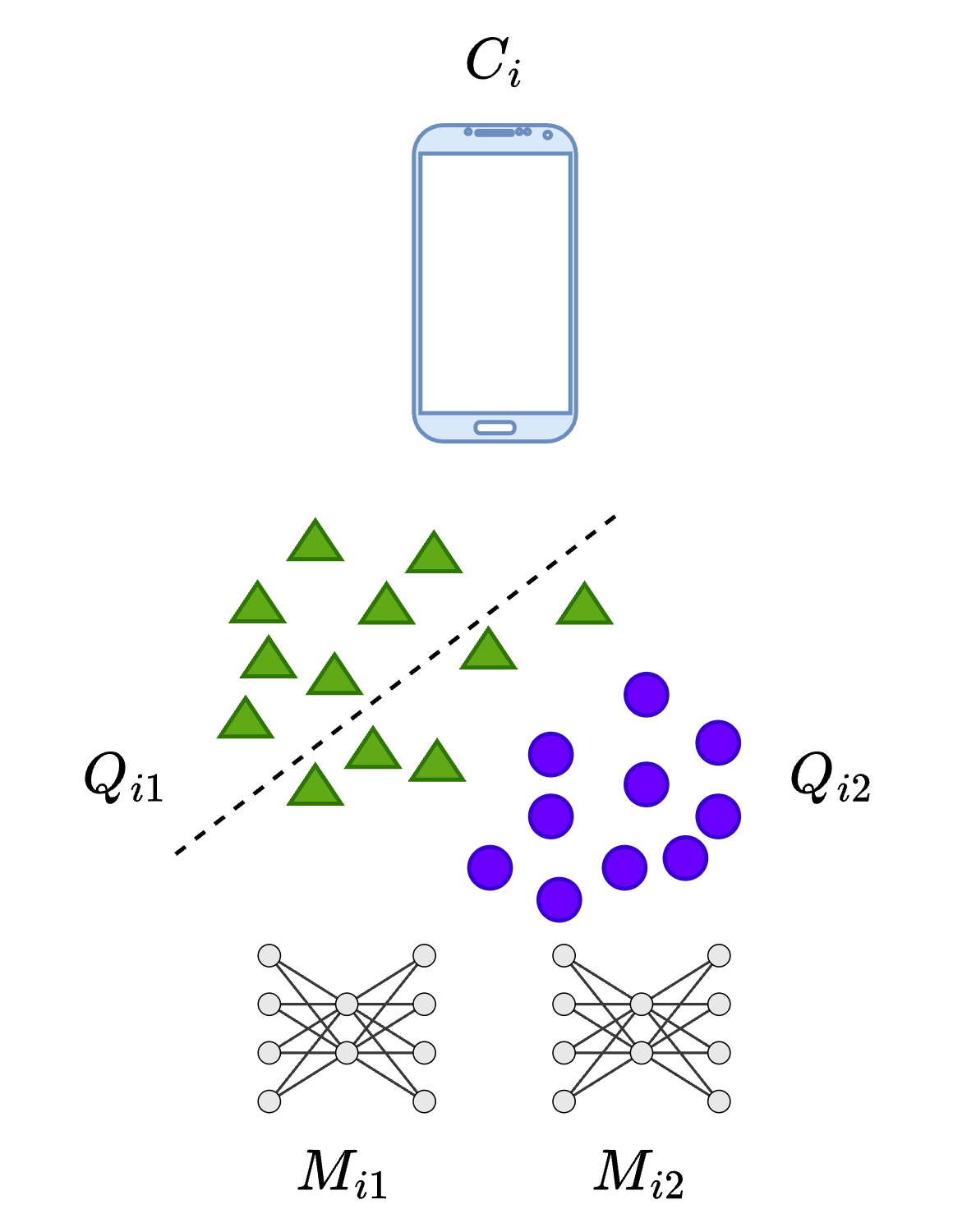}
    \caption{Illustration of Step 1 for client $C_i$ with two clusters ($K_i=2$). The dotted line indicates the initial clustering split of the local dataset into $Q_{i1}$ and $Q_{i2}$, which may be imperfectly aligned to the true categories. In this step, a representation model $M$ is trained locally for each cluster.}
    \label{fig:step1}
\end{figure}

To learn representations of local clusters, the models we employ are autoencoders. We denote as $M_{iq}$ the model trained for cluster $Q_{iq}$ of client $C_i$. Each client independently trains one model per cluster in its dataset $D_i$ (line~\ref{alg1:line:local_train}).
The training objective of $M_{iq}$ is to minimize the reconstruction error for each sample $x_k$ within cluster $Q_{iq}$ of client $C_i$.
This error is defined as $$e_k = l(x_k, \hat{x}_k)$$ where $l(\cdot)$ is the loss function (e.g., mean squared error), and $x_k$ and $\hat{x}_k$ represent the original sample from cluster $Q_{iq}$ and its reconstruction, respectively.

After the training, each client produces $K_i$ models, each specialized to represent a distinct local cluster. These models are then compared across clients to identify clusters with common latent characteristics, as described in the subsequent steps.  


\subsection*{Phase 2: Model Exchange and Local Cluster Association}
\label{subsec:association}

In this phase, clients exchange their trained models to discover which local clusters represent similar data distributions (lines~\ref{alg1:line:step:2:start}-\ref{alg1:line:step:2:end}). The key insight is: if two clusters truly share the same underlying distribution, then models trained on either cluster should reconstruct data from both clusters with similar accuracy.

Each client pair exchanges models and tests them on their local data. For example, client $C_i$ sends its model $M_{iq}$ (trained on cluster $Q_{iq}$) to client $C_j$, and receives $C_j$'s model $M_{jp}$ in return. Both clients then evaluate whether the external model reconstructs their data as well as their own model does.

\begin{figure}
    \centering
    \includegraphics[width=0.85\linewidth]{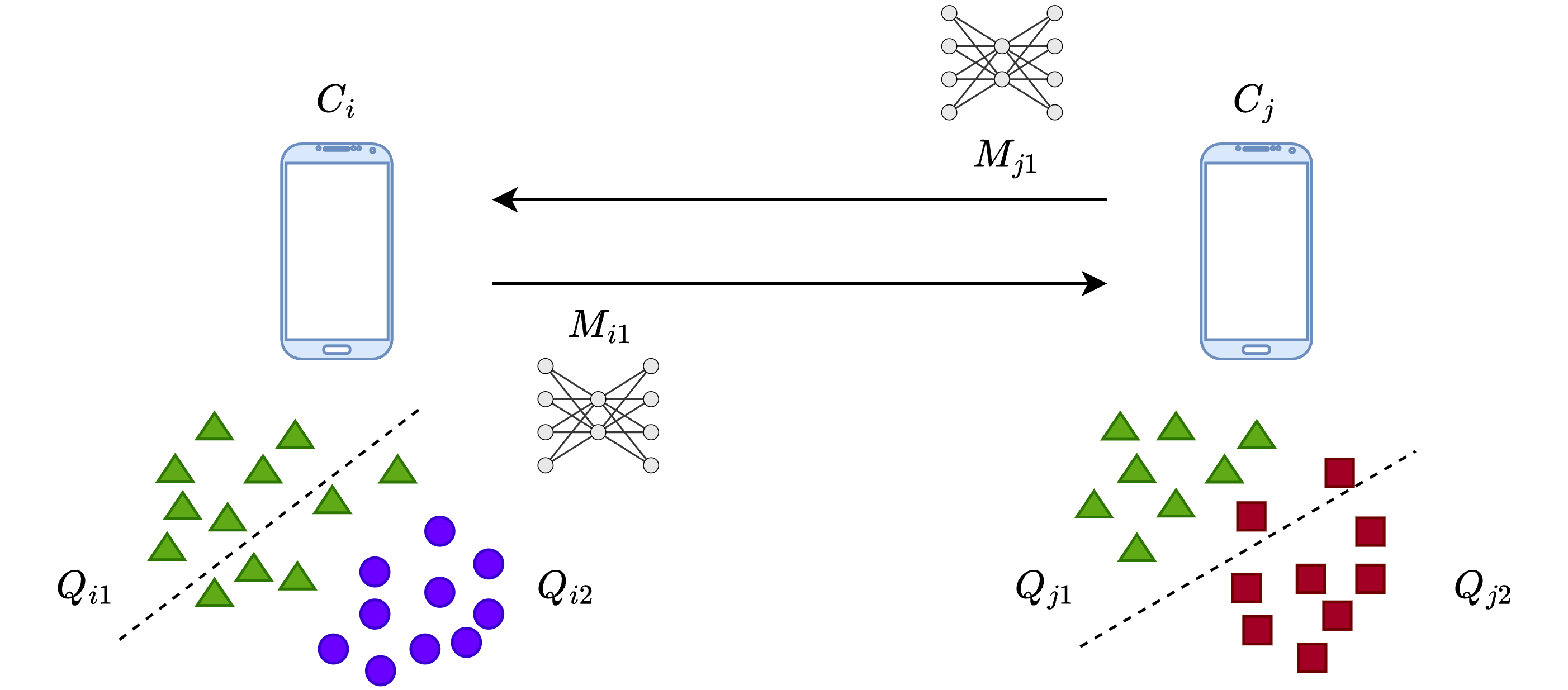}
    \caption{Diagram illustrating the model exchange mechanism: Client $ C_i $ assesses the reconstruction effectiveness of a model $ M_{jp} $ received from Client $ C_j $, and vice versa, to establish if clusters $ Q_{iq} $ and $ Q_{jp} $ indicate similar underlying distributions.}
    \label{fig:model_exchange}
\end{figure}

Client $C_i$ associates its cluster $Q_{iq}$ with $C_j$'s cluster $Q_{jp}$ through the following process (Figure~\ref{fig:model_exchange}):

\noindent\paragraph{Step 1: Compute reconstruction errors:} For each sample $x_k$ in cluster $Q_{iq}$, compute the error using both the local model ($e_k = l(x_k, M_{iq}(x_k))$) and the external model ($e'_k = l(x_k, M_{jp}(x_k))$), where $l(\cdot)$ is a discrepancy function (e.g., MSE).
    
\noindent\paragraph{Step 2 Measure error differences:} Compute and normalize the element-wise differences: 

\begin{equation}
 D_{\text{norm}}= \text{Normalize}_{[0,1]}(|E_{iq \rightarrow iq} - E_{jp \rightarrow iq}|)   
\end{equation}
where $E_{iq \rightarrow iq}$ and $E_{jp \rightarrow iq}$ are the error vectors.
    
\noindent\paragraph{Step 3 Apply threshold test:} Associate the clusters if $Q_{\alpha}(D_{\text{norm}}) \leq \theta$, meaning that at least $\alpha$\% of samples have normalized error differences at or below $\theta$ (where $Q_{\alpha}(\cdot)$ denotes the $\alpha$-th percentile).


Importantly, association requires \textit{bidirectionality}: both $C_i$ and $C_j$ must satisfy the threshold criterion for their respective clusters. This mutual validation ensures that the similarity is not due to coincidental patterns or noise.

We normalize error differences to $[0,1]$ to make threshold selection dataset-independent. Typical values are $\alpha \in [75, 95]$ and $\theta \in [0.05, 0.25]$. Higher $\alpha$ values enforce stricter similarity by requiring more samples to have small error differences, while lower $\theta$ values permit only highly similar clusters to associate.



\subsection*{Phase 3: Federated Groups Identification and Training}

In this phase, described in lines~\ref{alg1:line:step:3:start}-\ref{alg1:line:step:3:end} of Algorithm~\ref{alg:fed_complete}, clients first share their associations and construct an undirected graph $\Gamma(V, E) $ where all clusters from all clients serve as vertices. An edge is added between two clusters if bidirectionally associated in the previous step.\footnote{For operational simplicity and efficiency, graph construction can be centralized to a single node.} 
Once built, this graph consists of connected components\footnote{A connected component is a group of nodes for which it is possible to traverse from any node to any other node in a finite number of steps.}, where each component ideally groups local clusters representing the same underlying data distribution, we aim to federate.

\begin{figure}
    \centering
    \includegraphics[width=1.0\linewidth]{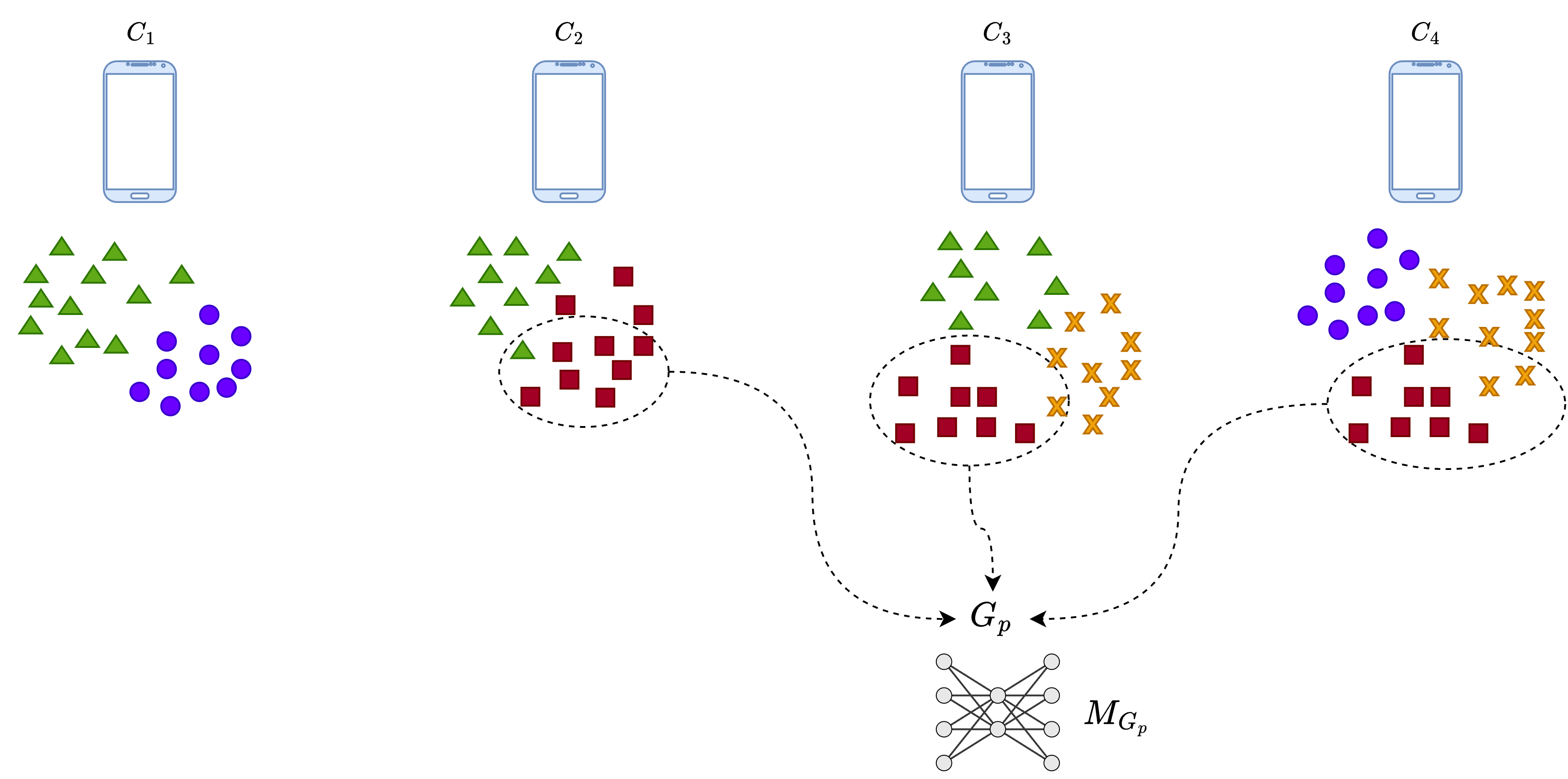}
    \caption{Diagram illustrating the identification and training of federated groups. In the example, $G_p$ consists of three associated clusters and a corresponding federated model$M_{G_p}$ is trained using the samples involved.}
    \label{fig:group_fl}
\end{figure}

Formally, let $\mathcal{G}$ denote the extracted set of connected components from $\Gamma$. Since single-node components do not contribute to the federated process, we define the \textit{communities} (or groups) of clusters $G_p$ as:

\begin{equation}
G = \{G_p \in \mathcal{G} : |V_{G_p}| \geq 2 \}  
\end{equation}

hence, $G$ is the set of all connected components containing at least two clusters. An example of an ideal connected component we aim to find and federate is illustrated in Fig.~\ref{fig:group_fl}.

For each community of clusters $G_p$, the clients involved initiate an unsupervised federated learning process training a common model $M_{G_p}$, in order to capture the representations data (Figure~\ref{fig:group_fl}). 
This process enables the model to learn a more comprehensive representation of the underlying data distribution by aggregating information from a larger set of samples across clients.

The federated learning protocol applied here does not specify a particular aggregator, allowing any participating client --typically the one with the lowest ID-- to serve as the aggregator.

In this process, an important consideration is the potential imprecision of local clustering. Since local clusters may not perfectly align with the true underlying data distributions (e.g., the shapes in Fig.~\ref{fig:group_fl}), two types of errors can arise during the association phase:

\begin{definition}[Wrong Association]
A wrong association occurs when a cluster is incorrectly linked to another belonging to a different underlying data distribution. We denote by $W \subseteq E$ the set of such associations. 

\end{definition}

\begin{definition}[Missed Association]
A missed association occurs when clusters that share the same underlying data distribution fail to be grouped together. 
\end{definition}

Missed associations can cause samples from the same underlying data distribution to be fragmented into separate groups, or lead to a local cluster remaining isolated despite belonging to a larger shared distribution. We denote by $I \subseteq V $ the set of all clusters having zero associations.

We address these issues through \textit{iterative cluster refinement}, which is described in detail in the next step.
The federated models help progressively improve the alignment between clusters and true categories, leading to more accurate associations at each iteration.

A relevant outcome of this phase is its ability to naturally handle poorly aligned clusters that fail to associate with any group. For instance, if a cluster exhibits significant impurity, indicating a weak correspondence to a true category, it is unlikely to form associations and is excluded from the process. This exclusion preserves the integrity of federated learning by preventing misleading data from causing incorrect mergers between groups representing different data distributions.

However, these isolated clusters are not permanently discarded. Instead, they are reassessed in subsequent refinement steps, where they may be recovered and reintegrated into the appropriate federated groups.

\subsection*{Phase 4: Federated model sharing and Clusters Refinements}

\begin{figure}
    \centering
    \includegraphics[width=1.0\linewidth]{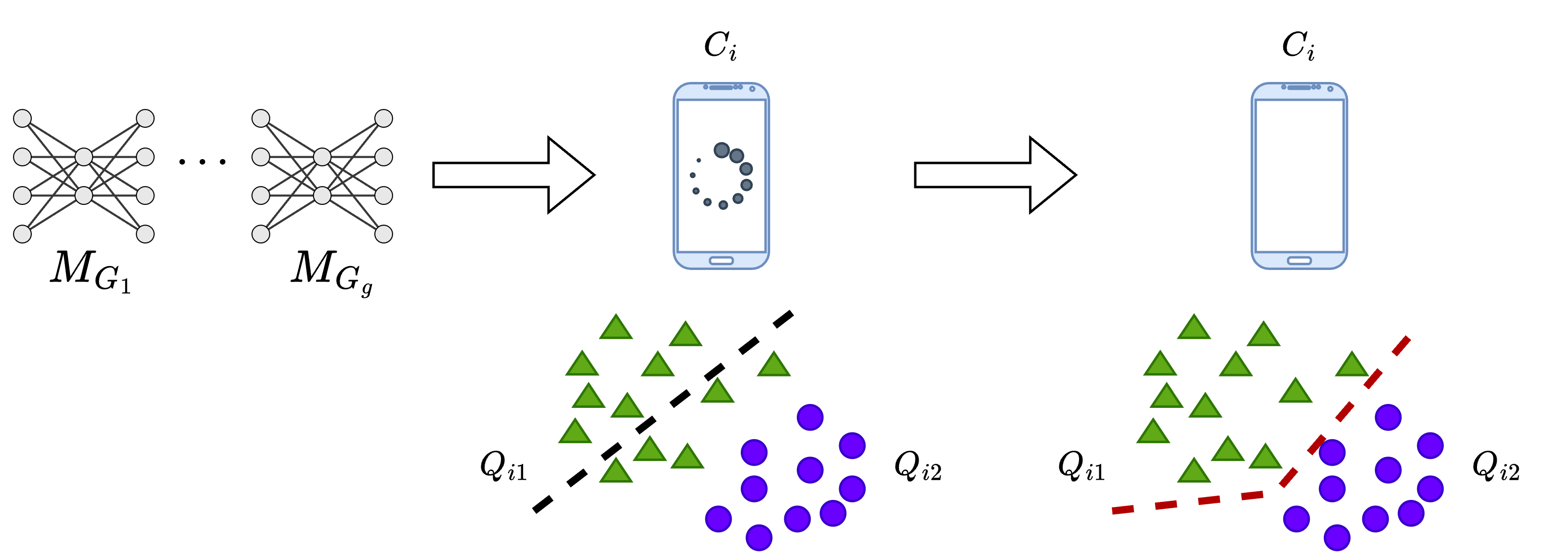}
    \caption{Illustration of cluster refinement for client $C_i$: The client receives all the trained federated models and use a subroutine to refine its local clustering ($K_i=2$), i.e., to achieve better alignment with the true categories.}
    \label{fig:client-repair}
\end{figure}

This phase covers lines~\ref{alg1:line:step:4:start}-\ref{alg1:line:step:4:end} of Algorithm~\ref{alg:fed_complete}.  Here, the trained federated models are distributed to all participating clients. The primary objective is for each client $C_i$ to refine their local clusters by leveraging both the received federated models $\{M_{G_g}, g=1,2, \dots,|G|\}$ and its own locally trained models  $\{M_{iq}, q=1,2,\dots,K_i \}$.  Practically, this process results in a new assignment of each data point in the local dataset into  $K_i$  distinct clusters (Fig~\ref{fig:client-repair}).  Cluster refinement is performed independently by each client as a dedicated subroutine (Algorithm~\ref{alg:clust_refine}).

At the start of this algorithm, all local samples in the dataset $D_i$ of client $C_i$ are considered unassigned, forming the local set $R$. As the algorithm progresses, points are assigned to clusters and subsequently removed from $R$. The process continues until $R$ is empty, ensuring that all samples have been (re)clustered. The steps involved are as follows:

    \noindent \paragraph{Step 1: Models Evaluation} For each data point $x_k$ in $R$ of client $C_i$, compute the reconstruction error using all available models. The model set $M$ includes both the local models $M_{iq}$ (one for each local cluster $Q_{i_q}$) and the received federated models $M_{G_g}$. The reconstruction error for $x_k$ under model $M_j$ is denoted as $e_{k, M_j}$ (line~\ref{alg2:line:all_errors} of Alg.~\ref{alg:clust_refine})
    \noindent \paragraph{Step 2: Best Model Selection} For each data point $x_k$ in $R$, identify the model that yields the lowest reconstruction error (line~\ref{alg2:line:select_best} of Alg.~\ref{alg:clust_refine}), defined as:
    \begin{equation}
    M^*(x_k) = \argmin_{M_j} e_{k, M_j}
    \end{equation}
    \noindent \paragraph{Step 3: Cluster Assignment} Identify the model \( M_{\text{max}} \) that is selected as \( M^*(x_k) \) for the largest number of unassigned data points \( x_k \in R \), i.e., the model most frequently chosen as optimal:
    \begin{equation}
    M_{\text{max}} = \argmax_{M_j} \; \text{count} ( M_j = M^*(x_k) \; | \; x_k \in R )
    \end{equation}
    Assign the corresponding data points to a new cluster $H_p$ (line~\ref{alg2:line:assign_sample} of Alg.~\ref{alg:clust_refine}). Remove them from $R$, and exclude the corresponding model from further iterations.
    \noindent \paragraph{Step 4: Iterative Refinement for $K_i$ Clusters} Repeat the steps of model evaluation, error comparison, and cluster assignment until $K_i$ clusters are formed, with the corresponding final model set denoted as $M_{select}$. 
    \noindent \paragraph{Step 5: Final Assignment of Remaining Samples} 
    During the previous iterative steps (1–4), only the data points best explained by a selected model are assigned in each iteration. As a result, some samples in $R$ may remain unassigned after $K_i$ clusters have been created and their corresponding models selected (i.e., the final model set $M_{select}$).
    To ensure every point is assigned, each remaining sample is matched to the model in $M_{select}$ that produces the lowest reconstruction error (line~\ref{alg2:line:remaining} of Algorithm~\ref{alg:clust_refine}). This ensures all local data points are ultimately assigned to one of the final clusters.

Note that the $K_i$ local models $M_{iq}$ serve as a safeguard, ensuring that if the received federated models do not generalize well to the local data, the clustering structure remains largely unchanged. In cases where a client’s data does not exhibit the patterns learned by the federated models, the local models are more likely to minimize reconstruction error, thereby maintaining the previous partitioning.

\begin{algorithm}
\footnotesize
\caption{Client $C_i$ Local Clusters Refinement Subroutine}
\label{alg:clust_refine}
\begin{algorithmic}[1]
\Procedure {ClusterRefine}{$C_i$, Received federated Models $M_{G}$}
\State Let $ H=\{ H_1 \ldots H_{K_i} \}$ be the set of $K_i$ empty clusters for client $C_i$
\State Let $R = D_i$ be the set of local unassigned samples 
\State Let $M = \{ M_{iq}, q=1,2,\dots,K_i \} \cup \{M_{G_g},
g=1,2, \dots,|G|\}$ 
\State Let $M_{select} = \emptyset $ be the models that best define the new refined clusters.
\For{$p = 1$ to $K_i$}
    \State Compute $e_{k, M_j}$ for all models $M_j \in M$ and all samples $x_k \in R$ \label{alg2:line:all_errors}
    \State Set $M^*(x_k) = \arg\min_{M_j} e_{k, M_j}$ for each $x_k$ \label{alg2:line:select_best}
    \State Set $M_{\text{max}} = \argmax_{M_j} \; \text{count} ( M_j = M^*(x_k) \; | \; x_k \in R )$ \label{alg2:line:select_max}
    

    \State $H_p = \{x_k \mid M^*(x_k) = M_{max}\}$ \label{alg2:line:assign_sample} \Comment{p-th cluster is formed}
    \State $M_{select} = M_{select} \cup M_{max}$
    \State $M = M \setminus M_{max}$
    \State $R = R \setminus  H_p$ 
\EndFor
\For{each remaining unassigned $x_k \in R$}
    \State Assign $x_k$ to the cluster corresponding to the model in $M_{select}$ that yields the lowest reconstruction error. \label{alg2:line:remaining}
\EndFor
\State Set $H$ as the new local cluster partitioning of the local dataset $D_i$
\EndProcedure
\end{algorithmic}
\end{algorithm}

\subsection*{Phase 5: System Convergence and Stopping Conditions}

After clients refine their local clusters using federated models, stopping conditions are evaluated to determine whether further refinement is necessary (line~\ref{alg1:line:convergence} of Algorithm~\ref{alg:fed_complete}).

\paragraph{Client-Level Stability Check}
A client is considered \textit{stable} if its local cluster assignments show \textit{no significant deviation} from the previous iteration. Since clustering algorithms assign arbitrary labels, Clustering Accuracy (ACC) can be used to find the optimal label correspondence before computing the accuracy.
Given two different clustering assignments $y$ and $y'$, it is given by: 

\begin{equation}
\text{ACC}(y', y) = \max_{\pi \in \Pi} \frac{1}{n} \sum_{i=1}^n 1\{ y'_i = \pi(y_i) \}
\label{eq:acc}
\end{equation}

where $\Pi$ represents all possible label permutations and the Hungarian algorithm~\citep{Kuhn1955Mar} efficiently finds the optimal $\pi$. For example, with $y' = [0, 0, 1, 1, 2]$ and $y = [2, 2, 1, 1, 0]$, optimal permutation $\pi(0)=2, \pi(1)=1, \pi(2)=0$ yields ACC = 1.0 despite different labels.

While ACC is commonly used to evaluate clustering against ground-truth labels~\citep{xie2016unsupervised},
we apply it also at this point to measure local stability of each client $C_i$ by comparing cluster assignments over previous ($y'$) and current ($y$) iteration. Specifically, we define the set of active clients as:

\begin{equation*}
    A = \{C_i: \text{ACC}(y'_{C_i}, y_{C_i}) < \tau ~ \forall C_i \in C \}
\end{equation*}

A client is removed from the active set $A$ once its ACC exceeds a predefined threshold $\tau$, indicating that the difference between consecutive clustering iterations is minimal and the client’s clusters have stabilized.

Stabilized clients stop training and refining their clusters (lines~\ref{alg1:line:local_train} and \ref{alg1:line:cluster_refine}) but continue participating in model exchange (Phase 2), ensuring their samples remain available for federated learning. Notably, stabilized clients with incorrect partitioning (i.e., misaligned with the true categories) naturally establish fewer new associations, limiting the propagation of misaligned clusters.

\paragraph{Global-Level Stopping Condition}
In addition to the other condition, based on the active clients in the process, we monitor the trends in the number of isolated clusters $|I|$ and the formation of communities $G$ within the system. The process is halted if these metrics remain stable over a predetermined number of iterations, indicating no significant changes. In our experiment, we specifically use the criteria that when both the number of communities and the number of isolated clusters do not change by more than 10\% over three consecutive iterations, the process is stopped. This ensures that the process halts when further iterations are unlikely to yield significant changes.

The process concludes when either no active clients remain in set $A$ or the global stopping condition is satisfied.

%% file: 05_experiments.tex
\section{Experimental setting}
\label{sec:experiments}

In this section, we present a series of experiments designed to evaluate the effectiveness of our methodology. 

\subsection{Dataset}
\label{subsec:fc:dataset}

Our study utilizes the EMNIST \citep{cohen2017emnist}, Fashion-MNIST \cite{xiao2017online}, KMNIST, and KMNIST49 datasets \citep{Clanuwat2018Dec}. The true classes in these datasets serve as the ground-truth data distributions that define the global set of categories $U$.

EMNIST Digits is an extended version of the classic MNIST dataset, containing 280,000 grayscale images of handwritten digits across 10 classes (0–9), each sized 28×28 pixels. Compared to MNIST, it offers a larger and more diverse set of samples, making it well-suited for evaluating our system in scenarios involving numerous clients while maintaining a familiar structure.

KMNIST (Kuzushiji-MNIST) consists of 70,000 grayscale images of handwritten Japanese characters on 10 classes, structurally similar to MNIST digits. This dataset presents a distinct pattern recognition challenge due to the differences between Japanese script and Arabic numerals. By including KMNIST, we assess the algorithm’s adaptability and effectiveness in handling visually different yet structurally comparable data.

Fashion-MNIST is a 70,000-image dataset of 28×28 grayscale clothing items in 10 categories (e.g., shirt, shoe, bag). It’s a drop-in replacement for MNIST, offering a more challenging benchmark for image classification models.

KMNIST49 (Kuzushiji-49) extends this challenge by incorporating a larger and more complex dataset with 270,912 images across 49 classes, each representing a different Japanese character. Like the previous datasets, it consists of 28×28 grayscale images. However, KMNIST49 is inherently imbalanced, with some classes containing very few samples. To ensure a balanced, disjoint partition, we remove underrepresented classes, resulting in a dataset with $|U| = 31$ classes, each containing approximately 6,000 samples.

The larger data volume of EMNIST and the expanded class count of KMNIST49  allow us to evaluate the scalability and accuracy of our approach in more demanding classification scenarios. 

Since this is an unsupervised task, actual labels are used only in two specific phases: (i) Initialization of the experiment, where they define the setup of the simulated system and determine the distribution of samples among clients. (ii) Evaluation, where they serve as a ground truth to assess the accuracy of the clusters formed by each client after the algorithm terminates (Section~\ref{sec:fc:performance}).

\subsection{Experimental setup and hyperparameters}
\label{subsec:setup}

We have designed a system setup that allows us to initialize experiments based on a specified client configuration. If not explicitly specified in the experiments, default values are assumed (Table~\ref{tab:default_params}). The key parameters of this setup include:

\begin{description}[style=unboxed,leftmargin=0cm]

  \item [Number of Clients] ($N$): A variable number of clients is configured for each experiment to examine the algorithm's performance under different network sizes. Default value is $N=25$.

  \item [Global Dataset] ($D$): The dataset used to initialize the system is selected from those described in Section~\ref{subsec:fc:dataset}. It defines the target number of data distributions ($K_G$ or $|U|$) that the algorithm aims to identify.

  \item [Number of Data Distributions per Client] ($K_i$): Each client $C_i$ is assigned a subset of data distributions, with $K_i$ ranging from 2 to $K_G/2$, where $K_G$ denotes the total number of distinct data distributions in the global dataset (e.g., $K_G=10$ for EMNIST). For instance, a client in EMNIST configuration may be assigned between 2 and 5 distributions (e.g., samples from classes 4, 5, and 8). Each assigned distribution is initialized with S unlabeled samples, ensuring heterogeneous data availability across clients, thereby simulating an unbalanced category distribution.
  Clients could estimate $K_i$ autonomously using clustering evaluation metrics such as the silhouette coefficient. However, for simplicity, we assume $K_i$ is provided to the client in our experiments.

  \item [Number of Samples per Data Distribution] ($S$): Each client is allocated 500 samples per data distribution. This allocation ensures an adequate data volume for effective local model training while still presenting a challenge due to limited data diversity.

  \item [Overlap Fraction] ($O$): This parameter measures the degree of shared data among client clusters introduced at system generation time. The default setting is 0 overlap.

\item [Percentile, Association, and Stability Thresholds] ($\alpha$, $\theta$, $\tau$): $\alpha$ defines the percentile criterion for cluster similarity, $\theta$ defines the error difference threshold for associations, and $\tau$ is the minimum stability required to consider a client’s clustering as converged, measured as the change in ACC between successive iterations. 
In our experiments, we adopt default values of $\alpha = 75$, $\theta = 0.2$, and $\tau = 0.8$, which fall within supported ranges reported in prior sections (see Section~\ref{subsec:association}). 
These values were experimentally selected after a grid search (Section~\ref{sec:exp:ablation}) over the ranges and were found to yield the best and most consistent performance across datasets and configurations.
We highlight that the use of the same default parameters across all experiments further demonstrates the robustness of the proposed methodology.

\end{description}

\begin{table}[htbp]
\centering
\caption{Default Experimental Parameters}
\label{tab:default_params}
\begin{adjustbox}{max width=\columnwidth}   
\begin{tabular}{ll}
\toprule
\textbf{Parameter} & \textbf{Default Value} \\
\midrule
Num. of Clients ($N$) & 25 \\
Data Distributions per Client ($K_i$) & Random in $[2, K_G/2]$; \\
Samples per Data Distribution ($S$) & 500 unlabeled samples \\
Overlap Fraction ($O$) & 0 (no overlap) \\
Percentile Threshold ($\alpha$) & 75 \\
Association Threshold ($\theta$) & 0.2 \\
Stability Threshold ($\tau$) & 0.8 \\
\bottomrule
\end{tabular} 
\end{adjustbox}
\end{table}

\subsection{Models architecture and training setting}
  To exact compressed representation from high-dimensional data, such as MNIST-like datasets, we employ an error-based neural network model, specifically a fully connected autoencoder, for both local and federated training.


We adopt an autoencoder architecture with an encoder of size 
$d$–100–64–32, where $d=784$ is the flattened $28 \times 28$ input, and a symmetrical decoder 32–64–100–$d$. The latent space has dimension 32. 
Deeper or wider networks were also evaluated, but provided no significant reconstruction improvements while increasing training time.
The final setup is summarized in Table \ref{tab:training_setup}. Note that, in this work, we adopt shallow autoencoders based on several practical considerations. Specifically, we target edge computing scenarios in which devices often operate under limited computational resources and with relatively small local datasets, conditions that make shallow architectures particularly suitable and efficient. Nevertheless, our proposed framework is agnostic to the specific model used for local data reconstruction. In principle, other variants—such as SimCLR-style models or Variational autoencoder architectures—could be integrated, depending on the required latent space complexity and provided they yield meaningful data representations. A comprehensive exploration of these alternatives is left for future work.

\begin{table}[htbp]
\centering
\caption{Model training setup}
\label{tab:training_setup}
\begin{adjustbox}{max width=\columnwidth}   
\begin{tabular}{ll}
\toprule
 Parameter & \textbf{Settings} \\
\midrule
Input/Output & $n\_features = 784$, Output activation: sigmoid \\
Hidden Layers & Neurons: [100, 64, 32, 64, 100], Activation: ReLU, Dropout: 0 \\
Training & Optimizer: Adam, Loss: MSE, Epochs: 20 \\
Total parameters & 174,840 (encoder+decoder) \\
\bottomrule
\end{tabular}
\end{adjustbox}
\end{table}

  The federated learning step is conducted over 15 rounds, a number selected through validation experiments to balance convergence speed and computational constraints. Increasing the number of rounds beyond this provided marginal accuracy gains while increasing training costs. Model aggregation is performed using FedAvg~\citep{mcmahan_communication-efficient_2017}.

  \subsection{Initial Data Partitioning Strategies}
\begin{figure}
    \centering
    \includegraphics[width=.4\linewidth]{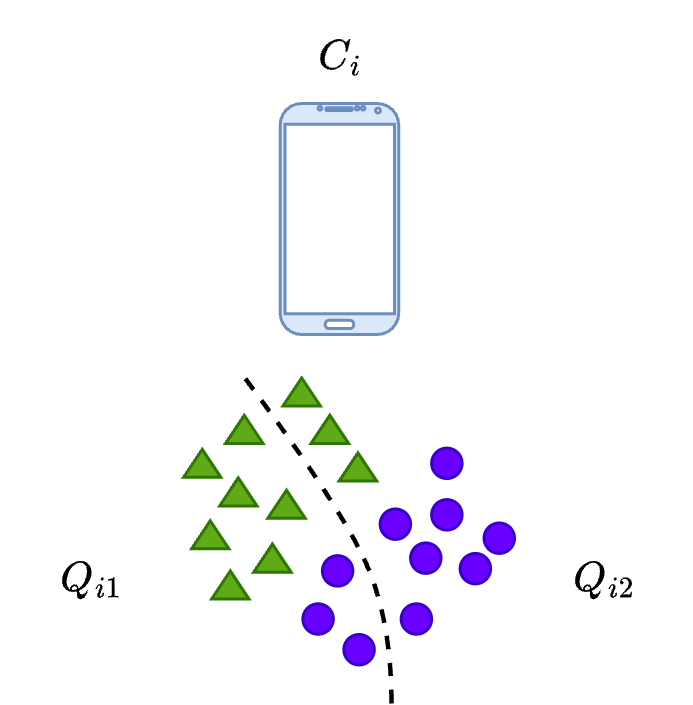}
    \caption{Example of client data split initialization with a dirtiness level of 0.3. In the initial clusters $Q_{i1}$ and $Q_{i2}$, 30\% of the samples is assigned to the wrong true category (represented by shapes).}
    \label{fig:client-dirtiness}
\end{figure}

    At the beginning of each experiment, we need an initial partitioning of client data. We select from two distinct approaches:

\begin{itemize}
    
    \item \textit{Dirty Uniform Clustering}: A simulated initialization where clusters are first initialized based on the true categories, but each sample is then randomly misassigned with a fixed probability (parameterized by \texttt{dirtiness} $ \in [0,0.5]$). As shown in the example in Fig~\ref{fig:client-dirtiness}, this controlled noise allows us to evaluate the algorithm’s resilience to initial clustering errors and its ability to correct them over iterations.

    \item \textit{Deep Embedded Clustering} (DEC)~\citep{xie2016unsupervised}: A realistic clustering approach where each client independently clusters its local data using a state-of-the-art deep learning-based method. For the configuration of DEC, we follow the original paper's specifications, using an encoder structure of $d-500-500-2000-10$, as it was specifically designed for clustering tasks and validated on datasets similar to ours. 
    
\end{itemize}

    The simulated approach enables precise control over initial clustering split, while DEC ensures evaluation within a realistic end-to-end pipeline. To validate the robustness of our approach, we conduct experiments with both methods.

\subsection{Performance Metrics}
\label{sec:fc:performance}

We monitor several key metrics to evaluate the effectiveness of the process.  The primary objective is to achieve optimal alignment between the discovered communities $G$ and the true global categories $U$, thereby validating the accuracy of our community detection methodology.

To assess the correctness of the discovered communities $G$, we track the number of isolated clusters $|I|$, which should ideally be zero under the assumption that each true distribution is shared by at least two clients, and the number of wrong associations $|W|$, which we aim to minimize—ideally reaching zero—to ensure accurate grouping of clusters.
These metrics are determined by comparing the cluster association graph against an ideal graph, where connections are established based on the true labels, i.e., information that is used exclusively for evaluation purposes.

Furthermore, since local cluster refinement plays a crucial role in improving associations and developing more precise federated models---by ensuring that samples more accurately reflect the correct data distribution---we track unsupervised clustering Accuracy (ACC), as defined in Eq.~\eqref{eq:acc}. Specifically, for each client, we compute: $\text{ACC}(y_t, y)$, where $ y_{t_i} $ are the ground truth labels and $ y_i $ are the clustering labels produced by the algorithm.

Unless otherwise stated, all results reported are averaged over five simulation runs, with confidence intervals computed at a $95\%$ confidence level.

%% file: 05_results.tex
\section{Results}
\label{sec:results}


In this section, we evaluate our proposed system across five key experiments. First, we show compare clustering quality against UIFCA and centralized clustering (Section~\ref{sec:exp:comparison}). 
Second, we analyze the impact of initial clustering noise (Section~\ref{sec:exp:dirtiness}). 
Third, we perform a sensitivity analysis on the key parameters controlling the cluster-wise clients associations (Section~\ref{sec:exp:ablation}).
Fourth, we examine how data overlap among clients affects performance (Section~\ref{sec:exp:overlap}). 
Finally, we evaluate scalability with respect to the number of clients (Section~\ref{sec:exp:numclients}).

\subsection{Overall Performance and Comparison against Centralized and UIFCA Baselines}
\label{sec:exp:comparison}


To evaluate the advantages of a decentralized approach, we compare \textit{FedCRef} with both an equivalent \textit{centralized} scenario and  \textit{UIFCA}~\cite{chung2022federated-uifca}, which represents the most related federated baseline in literature. (i) In the decentralized setup, the system is generated using the default settings from Section~\ref{subsec:setup}, utilizing the EMNIST Digits dataset. Local initial data splits are obtained using Deep Embedded Clustering (DEC) at each client, and the global data structure is progressively identified through our proposed \textit{FedCRef}. (ii) In the centralized setup, the clients splits are merged into a single dataset, simulating a scenario where all client data is transferred to a central server. In this case, DEC is applied globally, attempting to identify all data distributions in a single step. 
(iii) In the UIFCA setup, we re-implemented the framework under comparable conditions, adapting the configuration to our environment. 

\begin{figure}
    \centering
    \includegraphics[width=0.9\linewidth]{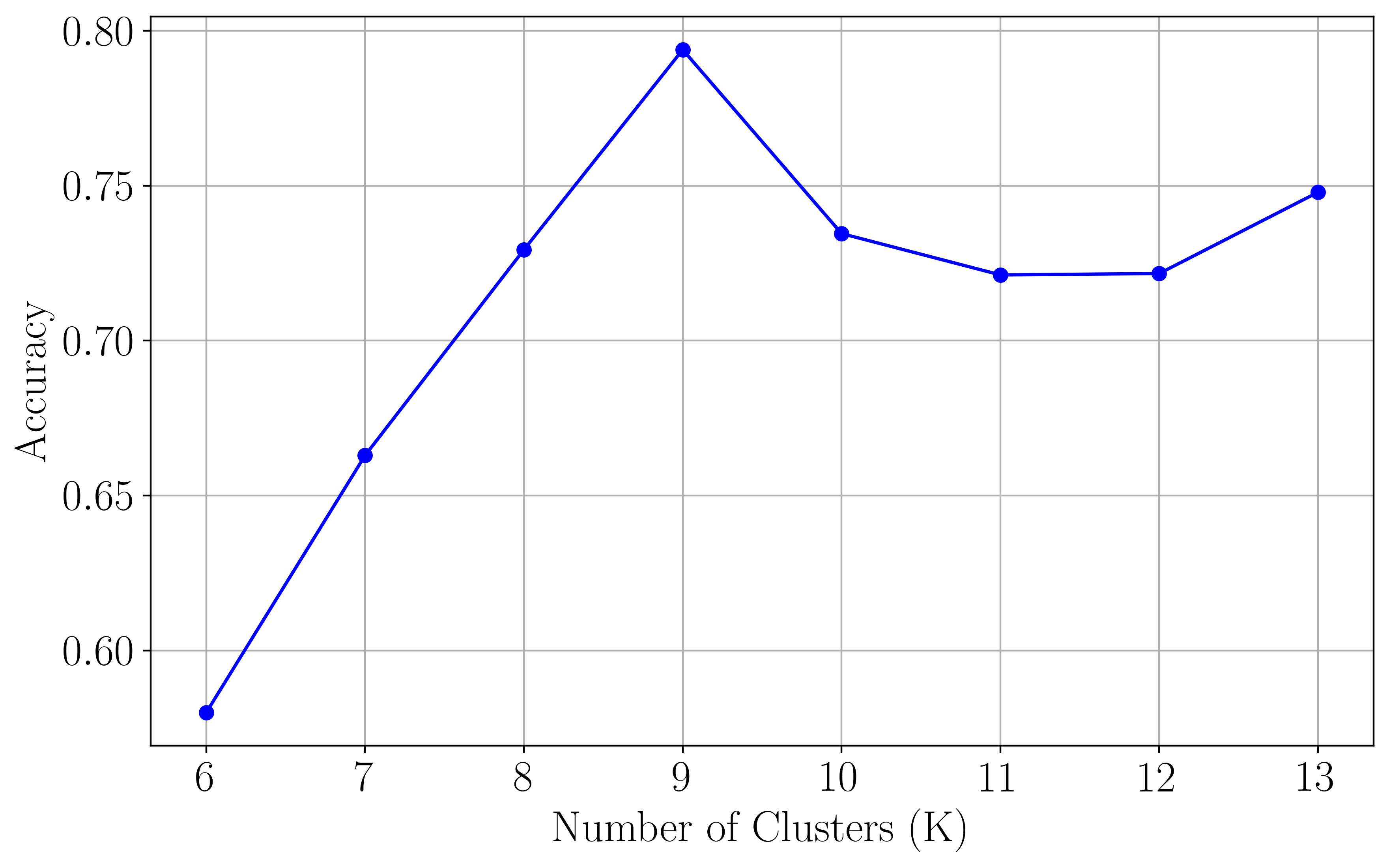}
    \caption{Accuracy evaluation of the Centralized setup using DEC clustering across different values of $K$}
    \label{fig:emnist:centralized}
\end{figure}

\noindent
\textit{Results against Centralized.} 
As presented in Table~\ref{tab:decent_vs_centr_vs_uifca}, FedCRef outperforms the centralized approach in terms of accuracy: FedCRef (Decentralized) achieves an ACC of 0.894, while Centralized DEC reaches ACC of 0.793 at its optimal choice of K.  The performance of the centralized approach is illustrated in detail in Fig.\ref{fig:emnist:centralized}. 
%
This substantial improvement highlights the difficulty of clustering in high-dimensional spaces when a single centralized model attempts to learn all distributions simultaneously. In contrast, in decentralized settings, local models are exposed to a subset of categories due to the non-IID nature of the data, effectively reducing the complexity of the clustering problem.

\begin{table}[htb]
\centering
\caption{Comparison of FsCRef vs. Centralized and UIFCA methods on the EMNIST dataset. Accuracy (ACC) is reported as the mean across five runs ($\pm 95\%$ confidence interval). The number of categories identified in the distributed setup represents the number of cluster groups to which FedCRef converged.}
\label{tab:decent_vs_centr_vs_uifca}
\begin{adjustbox}{max width=\columnwidth}   \begin{tabular}{lcc}
\toprule
   Method     &    ACC  &  Categories Identified /  \\
        &      &  True Categories $|U|$ \\
\midrule
  Centralized baseline w/ DEC    & $0.793 (\pm 0.087)$ & $9 (\pm 0.61) / 10$\\
                         FedCRef w/ DEC(ours)  & $0.894 (\pm 0.031)$ & $10.8/(\pm 1.1)/10$\\
                         UIFCA & $0.439 (\pm 0.021)$ & provided ($10$)  \\
\bottomrule
\end{tabular} \end{adjustbox}
\end{table}

\begin{figure*}[htbp]
    \centering
    \subfloat[][Number of Communities Found]{
    \includegraphics[width=.30\textwidth]{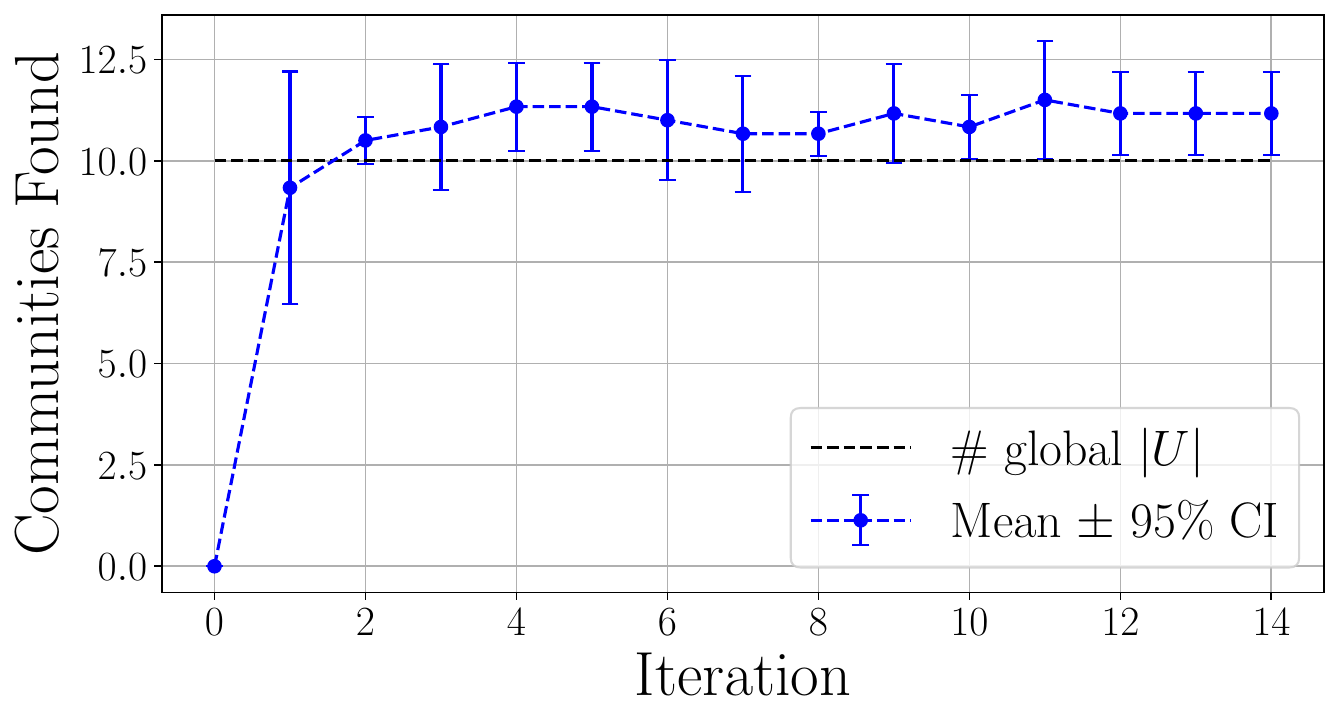}
    \label{fig:emnist:communities_found}
    }
    \subfloat[][Accuracy]{
    \includegraphics[width=.30\textwidth]{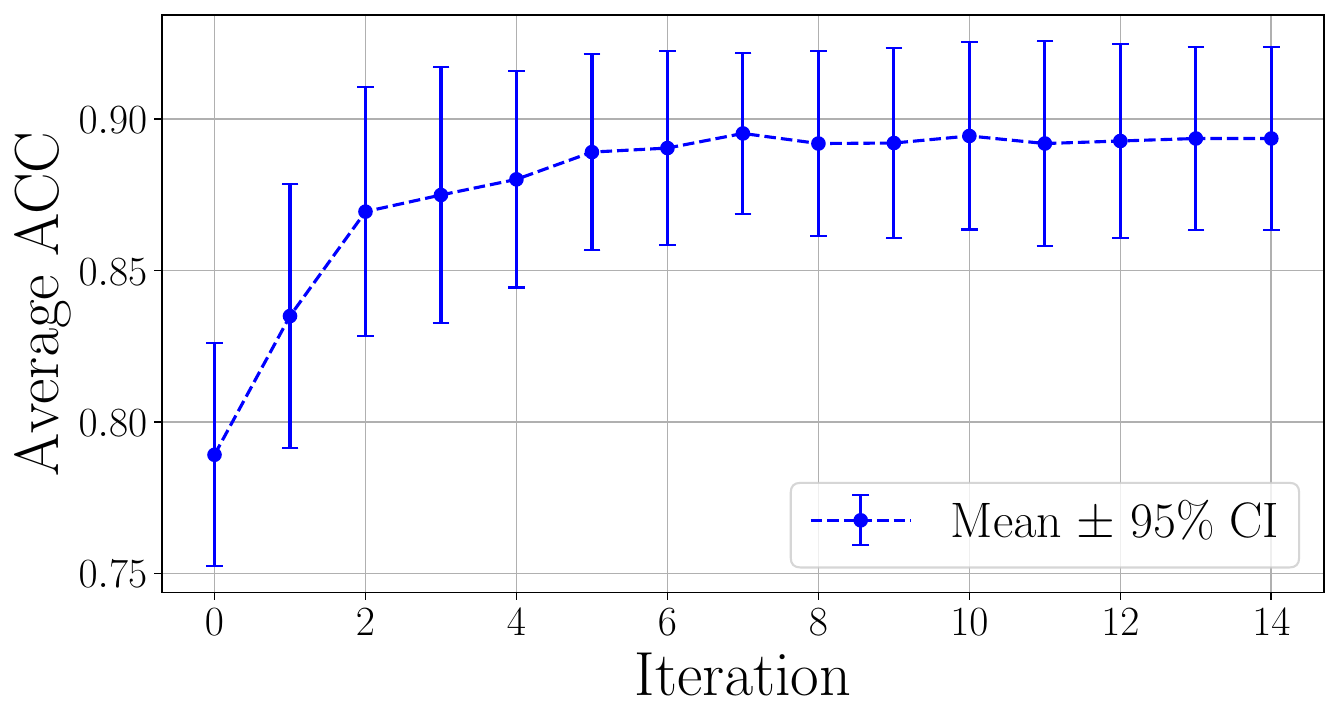}
    \label{fig:emnist:unsupervised_accuracy}
    }
    \subfloat[][Wrong Associations (\%) Between Clusters]{
    \includegraphics[width=.30\textwidth]{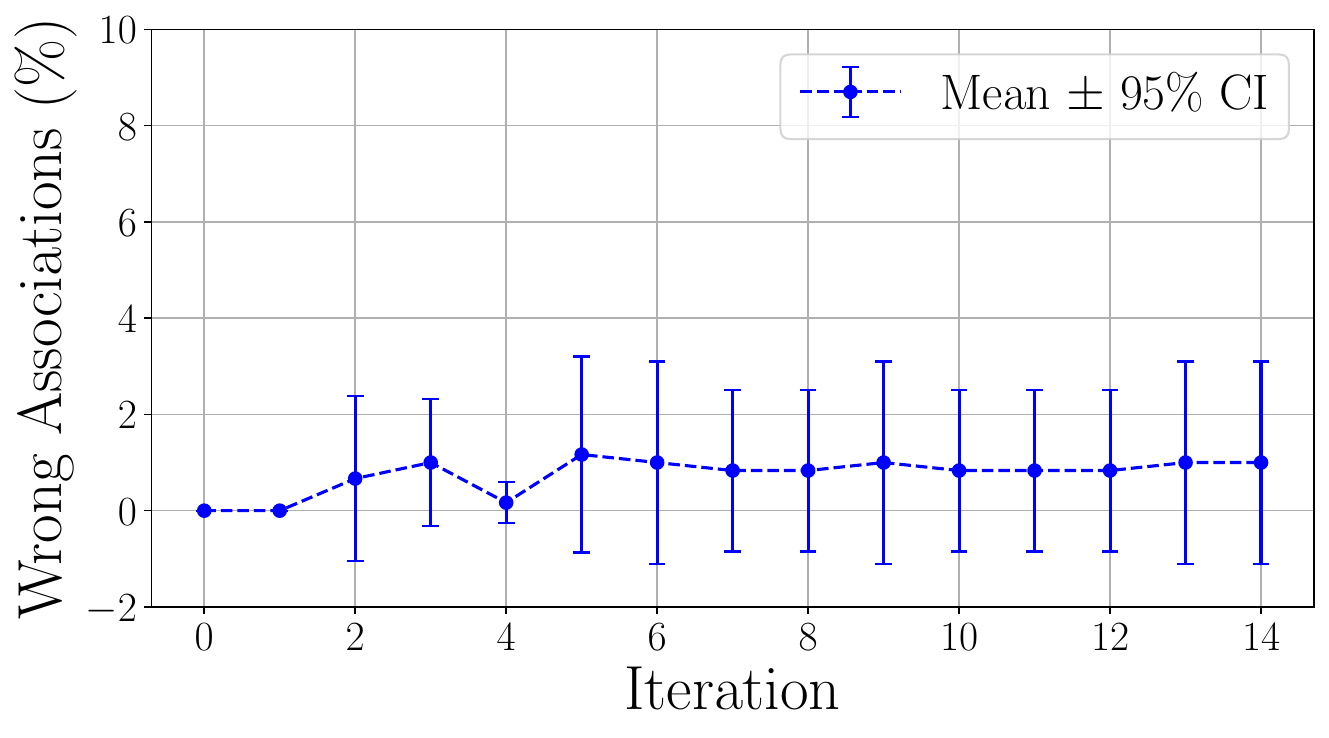}
    \label{fig:emnist:wrong_associations}
    }
    \\
    \subfloat[][Isolated Clusters]{
    \includegraphics[width=.30\textwidth]{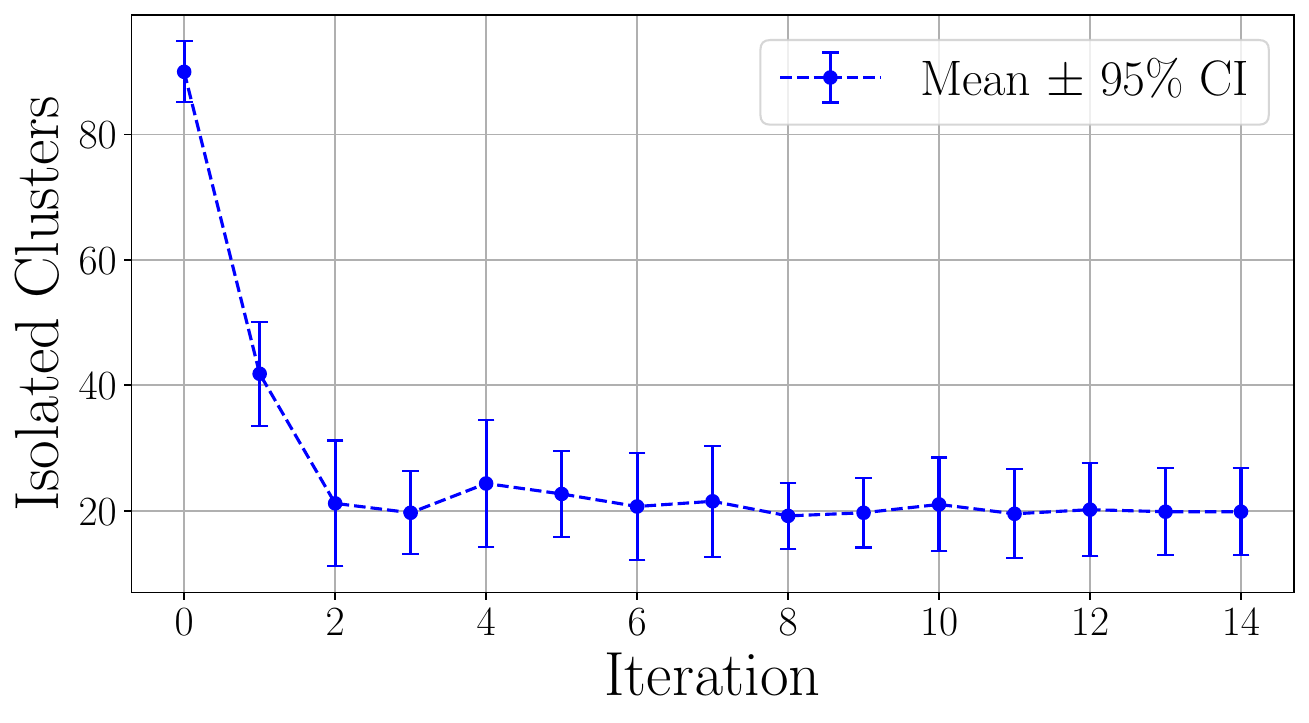}
    \label{fig:emnist:isolated_clusters}
    } 
    \subfloat[][Active Clients]{
    \label{fig:emnist:active_clients}
    \includegraphics[width=.30\textwidth]{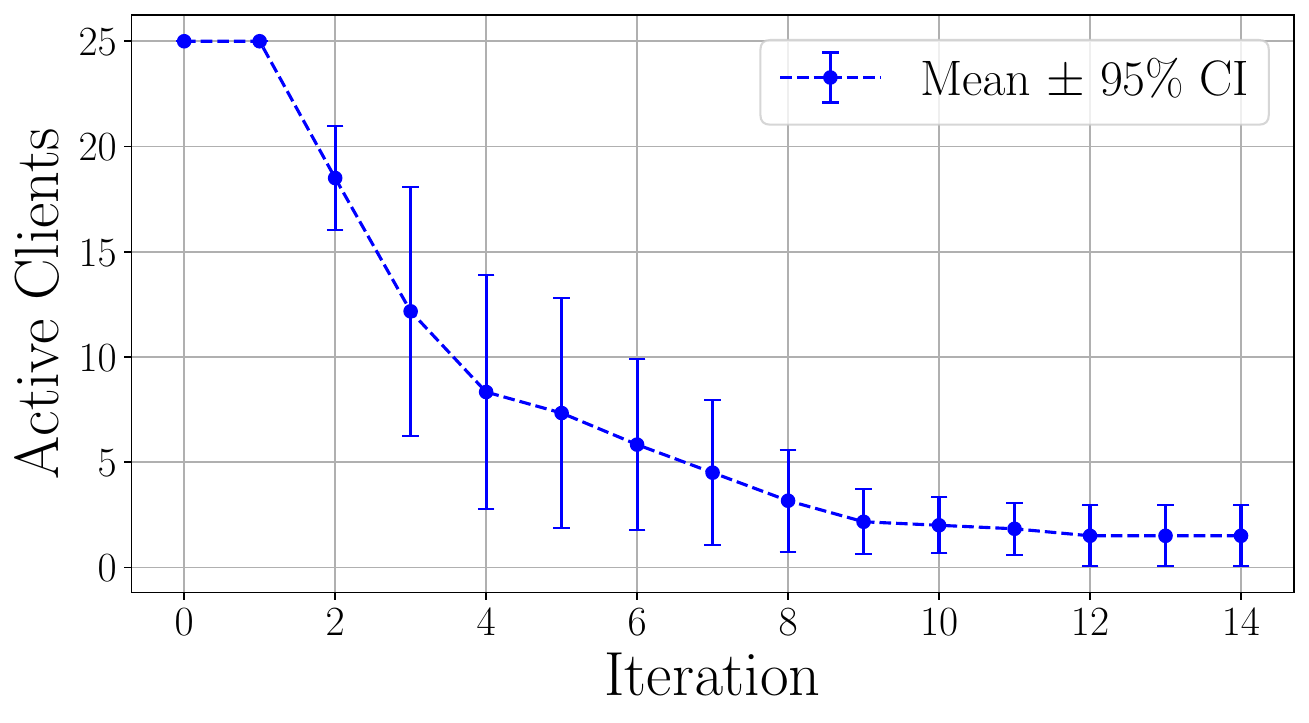}
    }
    \caption{Performance metrics of FedCRef over iterations on the EMNIST Digits dataset ($|U|$ = 10 data distributions). Values represent the mean $\pm 95\%$ confidence interval across five runs. Clients are initialized with a random subset of 2 to $|U|/2$ data distributions, yielding an average of 3.5 data distributions per client. The initial local clustering is performed using the DEC algorithm.}
    \label{fig:emnist:trends}
\end{figure*}

To further illustrate the behavior of FedCRef, Figure~\ref{fig:emnist:trends} provides a detailed breakdown of its performance on the EMNIST Digits dataset over multiple iterations. Final key metrics are also summarized in Table~\ref{tab:real_case_outcomes}.
%
The number of communities $|G|$ provides insight into how many underlying data distributions the process has identified. Ideally, this should align with the true number of global categories $|U|$. As shown in Figure~\ref{fig:emnist:communities_found}, FedCRef quickly stabilizes near the correct value, demonstrating its effectiveness in recovering the true data distributions.
The final accuracy (ACC) is the average accuracy across all clients, and  Figure~\ref{fig:emnist:unsupervised_accuracy} illustrates the progressive improvement in accuracy over iterations. Starting from an initial value of 0.78, FedCRef refines local clusters through federated knowledge sharing, ultimately reaching 0.894.
To verify that the community detection process is functioning correctly, we rely on additional metrics. First, the percentage of wrong associations should be close to zero, indicating that clusters are being correctly grouped. As shown in Figure~\ref{fig:emnist:wrong_associations}, FedCRef consistently achieves this.
Second, the number of isolated clusters should decrease over time. Figure~\ref{fig:emnist:isolated_clusters} shows that this number drops from an initial average of 87.5 (across 25 clients) to approximately 19, indicating effective community formation while still allowing isolated clusters to refine themselves through federated updates.
Finally, we consider the number of active clients over iterations, which provides an indication of convergence. As FedCRef refines clusters and approaches a stable community structure, fewer clients need to actively participate in refinement. Figure~\ref{fig:emnist:active_clients} illustrates this trend, where the number of active clients gradually decreases, signaling convergence as most clusters stabilize.

\begin{table}[htbp]
\centering
\caption{Clustering outcomes of our FedCRef with \textit{DEC} as the local data initialization split for clients (Mean $\pm$ 95\% CI) over five runs. Dataset: EMNIST.}
\label{tab:real_case_outcomes}
\begin{adjustbox}{max width=\columnwidth}  
\begin{tabular}{ccccc}
\toprule
 ACC init  & ACC End  & Comm. Found $|G|$ & Wrong Ass. & Isolated Cl. \\
         (Local DEC) & & / True Categories $|U|$ & (\%)  & Start $\rightarrow$ End \\
\midrule
 $0.781 (\pm 0.045) $ & $0.894 (\pm 0.031)$ &  $10.8 (\pm 1.1)$ / 10 & $ 1.1 (\pm 1.6)$ & $87.5(\pm 3.2)  \rightarrow 19.1 (\pm 7.2)$ \\
\bottomrule
\end{tabular} 
\end{adjustbox}
\end{table}

\noindent
\textit{Results against UIFCA.} To provide a direct reference against the most related approach, we re-implemented UIFCA~\cite{chung2022federated-uifca}, as no official implementation is publicly available. To align with the most comparable UIFCA experiment, we select the \textit{Cluster by Digits} setting, where each cluster corresponds to a digit class.

In our experiment, for consistency with our settings, we used the EMNIST Digits dataset and generate a federated partition with 25 clients (Table~\ref{tab:default_params}). To each client is assigned a disjoint subset of classes, resulting in a non-IID scenario. For the generative model, we implemented RealNVP, i.e., the model adopted in the UIFCA paper, with six affine coupling layers.\footnote{Each coupling layer uses a conditioner network with three fully-connected layers (784→512→512→512→784) with ReLU activations and L2 regularization ($\lambda=10^{-4}$), producing scale ($s$) and translation ($t$) parameters via tanh and linear output layers, respectively. The model uses checkerboard alternating binary masks and a standard Gaussian base distribution.} 
We set the optimizer to Adam (maintaining the learning rate at $\eta=10^{-4}$). All other UIFCA hyperparameters were kept consistent with the paper: $T=20$ cluster rounds, $\tau=40$ communication rounds per cluster round, and $M=100$ local gradient steps per communication round.

Table~\ref{tab:decent_vs_centr_vs_uifca} summarizes the obtained performance. In line with the findings reported in the original study, UIFCA struggles on MNIST digit clustering, and EMNIST presents an equivalent challenge. 
Our results confirm this behavior, yielding an average accuracy of $0.4394$, substantially below FedCRef. Notably, UIFCA operates under the significant advantage of knowing $K=10$ a priori, whereas FedCRef must estimate the number of clusters from the data.

\subsection{Impact of initial clustering precision} 
\label{sec:exp:dirtiness}

\textit{Experiment setup.} 
In this experiment, we evaluate FedCRef’s sensitivity to variations in the initial clustering accuracy of clients by employing a simulated clustering initialization, as described in Section~\ref{subsec:setup}, i.e., clients' samples are first partitioned based on the true categories, then each one is randomly misassigned to the wrong data distribution with a fixed probability, defined by the \texttt{dirtiness $\in [0, 0.5]$} parameter.

A dirtiness level of 0 serves as a best-case baseline, where clients clusters start perfectly aligned with the true categories (i.e., initial accuracy = 1.0). To examine FedCRef’s robustness, we introduce additional levels of dirtiness:
Dirtiness = 0.3: simulates a scenario where 30\% of samples are misassigned (e.g., Fig.~\ref{fig:client-dirtiness}), resulting in an initial clustering accuracy of 0.7. This level reflects a moderate amount of noise, slightly worse than what a typical clustering method like DEC would produce.
Dirtiness = 0.5: represents a highly unfavorable starting condition, where half of the samples are misassigned, leading to an initial clustering accuracy of 0.5. This extreme scenario tests the algorithm’s ability to recover from significant misalignment in the initial partitioning.
Unless otherwise specified, all other parameters are set to their default values.

\noindent
\textit{Results Analysis.} 
Table~\ref{tab:community_comparison} presents the clustering outcomes across different levels of initial dirtiness. Despite variations in the starting conditions, FedCRef consistently identifies the correct number of communities, aligning with the actual data distributions in both the EMNIST and KMNIST datasets.

Lower dirtiness levels result in fewer incorrect associations, indicating high correctness in community formation. Similarly, a notable reduction in isolated clusters suggests that the algorithm effectively consolidates clusters.

In the KMNIST-49 dataset (target $|U|$ = 31), the higher complexity of the data distributions makes complete detection more challenging. Nonetheless, the algorithm adopts a conservative approach, prioritizing correct associations over forced merges, which results in fewer incorrect associations but a higher number of isolated clusters.

\begin{table}[htbp] 
\centering
\caption{Clustering outcomes across different datasets (average values over five runs), with 95\% confidence intervals (CI).  The starting local partitioning is performed using \textit{simulated clustering}, with varying \texttt{dirtiness} levels to assess its impact on the results. }
\label{tab:community_comparison}

\begin{adjustbox}{max width=\columnwidth}   \begin{tabular}{lcccc}
\toprule
Dataset & Dirtiness & Comm. Found / & Wrong Ass. & Isolated Cl.\\
& & True Categories $|U|$ & (\%) & Start $\rightarrow$ End \\
\midrule
\multirow{3}{*}{EMNIST} 

& 0.0 & $10.1(\pm 0.7) $ / 10 & $0.0 (\pm 0.0) $ & $ 89.4(\pm 2.5)  \rightarrow  17.3 (\pm 3.9) $ \\
& 0.3 & $9.8 (\pm 1.1)$ / 10 & $5.1 (\pm 1.7)$ & $87.6 (\pm 4.2) \rightarrow 16.1 (\pm 1.3)$ \\
& 0.5 & $10.2 (\pm 0.9)$ / 10 & $6.9 (\pm 1.2)$ & $89.4 (\pm 5.6) \rightarrow 48.0 (\pm 3.2)$ \\
\midrule
\multirow{3}{*}{KMNIST}    

& 0.0 & $9.8(\pm 1.6) $ / 10 & $0.0 (\pm 0.0) $ & $ 87.0(\pm 6.7)  \rightarrow  18.6 (\pm 3.4) $ \\
& 0.3 & $9.8 (\pm 0.8)$ / 10 & $2.0 (\pm 1.1)$ & $88.1 (\pm 4.3) \rightarrow 19.2 (\pm 1.4)$ \\
& 0.5 & $10.8 (\pm 0.9)$ / 10 & $3.1 (\pm 0.9)$ & $86.4 (\pm 3.8) \rightarrow 31.0 (\pm 2.5)$ \\
\midrule
\multirow{3}{*}{KMNIST-49}    

        & 0.0 & $27.2(\pm 1.6) $ / 31 & $2.4 (\pm 2.3) $ & $ 201.4(\pm 11.5)  \rightarrow  40.0 (\pm 7.8) $ \\
        & 0.3 & $27.4 (\pm 2.6)$ / 31 & $7.4 (\pm 2.2)$ & $197.8 (\pm 14.5) \rightarrow 53.4 (\pm 4.3)$ \\
        & 0.5 & $24.8 (\pm 4.4)$ / 31 & $8.2 (\pm 3.0)$ & $218.6 (\pm 17.2) \rightarrow 89.8 (\pm 5.1)$ \\
\midrule
\multirow{3}{*}{FASHION-MNIST}    

        & 0.0 & $8.2(\pm 1.04) $ / 10 & $ 5.2 (\pm 2.3)$ & $87.2(\pm 3.7)  \rightarrow  5.8 (\pm 2.2) $ \\
        & 0.3 & $8.33 (\pm 2.87)$ / 10 & $4.0 (\pm 5.0)$ & $84.7 (\pm 5.7) \rightarrow 9.7 (\pm 7.6)$ \\
        & 0.5 & $7.67 (\pm 1.43)$ / 10 & $9.8 (\pm 3.3)$ & $85.6 (\pm 3.4) \rightarrow 16.0 (\pm 10.8)$ \\
\bottomrule
\end{tabular} \end{adjustbox}

\end{table}

Table~\ref{tab:uacc_improvement_comparison} summarizes the clients’ local accuracy (ACC) over iterations. As initial dirtiness increases, the final accuracy (ACC End) naturally declines across all datasets. However, this decline is still significantly smaller relative to the initial accuracy (ACC Init), highlighting the algorithm’s ability to recover from noisy initial partitions.

We also include a baseline (ACC \emph{Perfect Association})  where clients are provided with ground-truth cluster associations, eliminating errors from the association step. This baseline achieves very high accuracy, and our experimental results closely approximate this benchmark, demonstrating the robustness of our method.
The least variability in performance improvement with changing dirtiness levels is observed in KMNIST-49, suggesting that high number of data distributions in the dataset play a role in algorithmic stability.

To assess generality beyond digit datasets, we tested \emph{FedCRef} on Fashion-MNIST. This dataset notably exhibits higher intra-class variability and visually richer categories (e.g., shirts, coats, and dresses), which can increase the risk of incorrect associations across similar classes. 
We retained the same autoencoder architecture, while training for 30 epochs and slightly reducing the association threshold from $\theta=0.2$ to $\theta=0.15$, in order to enforce stricter matching and mitigate potential misalignments, while keeping all remaining settings as in the default configuration (Table~\ref{tab:default_params}).

The results are summarized in the last row of Tables~\ref{tab:community_comparison},\ref{tab:uacc_improvement_comparison}. As expected, distinctly recovering all ten categories is more challenging; however, FedCRef successfully identifies a number of communities close to the ground-truth categories, maintains a low rate of incorrect associations, and achieves competitive final accuracy. This confirms that the method extends to more complex datasets without architectural modifications.

\begin{table}[htbp] 
\centering
\caption{Accuracy across different starting \texttt{dirtiness} levels, using \textit{simulated clustering} as the initial local clustering method for clients. Each row represents the average accuracy over five runs per configuration, with 95\% confidence intervals (CI) included.} 
\label{tab:uacc_improvement_comparison}

\begin{adjustbox}{max width=\columnwidth}   \begin{tabular}{lccccc}
\toprule
Dataset & Dirtiness & ACC Init & ACC End & Improvement (\%) & ACC End  \\
 &  &  & (Our)  &   &  (Perfect Associations) \\
\midrule
\multirow{3}{*}{EMNIST}   
        & 0.0 & 1.0  & $ 0.948 (\pm 0.011) $ & $ -5.2 (\pm 1.05)$ & $0.982 (\pm 0.001) $\\
        & 0.3 & 0.701 & $0.879 (\pm 0.028)$ & $25.392 (\pm 4.00)$ & $0.949 (\pm 0.0097) $\\
        & 0.5 & 0.503 & $0.711 (\pm 0.049)$ & $41.352 (\pm 9.74)$ & $0.914 (\pm 0.0059) $\\
\midrule
\multirow{3}{*}{KMNIST}   
        & 0.0 & 1.0  & $ 0.93 (\pm 0.016) $ & $ -7.0 (\pm 1.6) $ & $0.963 (\pm 0.0025) $\\
        & 0.3 & 0.700 & $0.863 (\pm 0.031)$ & $23.286 (\pm 4.43)$ & $0.933 (\pm 0.0047) $ \\
        & 0.5 & 0.504 & $0.765 (\pm 0.058)$ & $51.786 (\pm 11.51)$ & $0.905 (\pm 0.0059) $ \\
\midrule
\multirow{3}{*}{KMNIST-49} 
        & 0.0 & 1.0  & $ 0.889 (\pm 0.008) $ & $ -11.1 (\pm 0.8) $ & $0.933 (\pm 0.0041) $\\
        & 0.3 & 0.699 & $0.769 (\pm 0.019)$ & $10.014 (\pm 2.72)$ &  $0.889 (\pm 0.0029) $ \\
        & 0.5 & 0.501 & $0.622 (\pm 0.065)$ & $24.152 (\pm 12.97)$ & $0.849 (\pm 0.0011) $  \\
\midrule
\multirow{3}{*}{FASHION-MNIST} 
        & 0.0 & 1.0   & $0.869 (\pm 0.015)$ & $-13.1 (\pm 1.51)$ & $ 0.864 (\pm 0.0026)$ \\
        & 0.3 & 0.699 & $0.787 (\pm 0.046)$ & $12.62 (\pm 6.6) $ &  $ 0.828 (\pm 0.0021)$ \\
        & 0.5 & 0.501 & $0.740 (\pm 0.087)$ & $47.7  (\pm 17.4)$ &  $ 0.808 (\pm 0.003)$ \\
\bottomrule
\end{tabular} \end{adjustbox}

\end{table}

\subsection{Sensitivity analysis on (\texorpdfstring{$\alpha$}{alpha}, \texorpdfstring{$\theta$}{theta})}
\label{sec:exp:ablation}

Table \ref{tab:ablation} reports the impact of varying the association threshold $\theta$ and the percentile parameter $\alpha$ in the EMNIST default setting (Table~\ref{tab:default_params}). The results show that both parameters influence the balance between correct associations, community formation, and overall clustering accuracy.

With a strict threshold ($\theta=0.1$), the number of incorrect associations is reduced to zero; however, the majority of clusters remain isolated and the final accuracy remains close to the initialization. In contrast, a relaxed threshold ($\theta=0.3$) decreases the number of isolated clusters but increases the fraction of incorrect associations, leading to a reduction in accuracy.

For the percentile parameter, setting $\alpha=90$ prevents any association, leaving the system unchanged. Lowering $\alpha$ to 85 allows some communities to be formed and moderately improves accuracy, although a substantial number of clusters remain isolated.

The configuration $(\alpha=75, \theta=0.2)$ provides the most balanced outcome, combining high final accuracy with recovery of the correct number of communities, a strong reduction in isolated clusters, and a limited proportion of incorrect associations. This configuration is therefore adopted as the default setting for FedCRef.

\begin{table}[htbp]
\centering
\caption{Sensitivity analysis on the association threshold $\theta$ and percentile $\alpha$. 
The best configuration $(\alpha=75, \theta=0.2)$ is reported and  highlighted for comparison.}
\label{tab:ablation}
\begin{adjustbox}{max width=\columnwidth}   
\begin{tabular}{cc|cccc}
\toprule
$\alpha$ & $\theta$ & Final ACC & Communities Found / & Wrong Assoc. & Isolated Clusters  \\
  &  & &  True Categories $|U|$ & (\%)  & (Start $\rightarrow$ End) \\
\midrule
\textbf{75} & \textbf{0.2} & \bm{$0.879(\pm 0.028)$} & \bm{$9.8(\pm 1.1)/10$} & \bm{$5.1(\pm 1.7)$} & \bm{$87.6(\pm 4.2)\rightarrow 16.1(\pm 1.3)$} \\
75 & 0.1 & $0.709(\pm 0.020$ & $1.6(\pm 1.1)/10$ & 0.0 & $84.8(\pm 5.6 \rightarrow\ 76.8(\pm 8.2)$ \\
75 & 0.3 & $0.644(\pm 0.024)$ & $12.2(\pm 1.0)/10$ & $12.4(\pm 2.3)$ & $85.2(\pm 4.8) \rightarrow 19.8(\pm 2.2)$ \\
85 & 0.2 & $0.752(\pm 0.027)$ & $6.8(\pm 1.0)/10$ & $0.8(\pm 1.0)$ & $87.3(\pm 5.4) \rightarrow 51.4 (\pm 5.6)$ \\
90 & 0.2 & $0.700$ & $0.0 / 10 $  & 0.0 & $87.6(\pm 5.4) \rightarrow\ 87.6(\pm 5.4)$ \\

\bottomrule
\end{tabular}
\end{adjustbox}
\end{table}

\subsection{Impact of data overlap}
\label{sec:exp:overlap}
 
\textit{Experiment Setup}.
This experiment investigates how data overlap among clients influences the performance of FedCRef. We evaluate three levels of overlap: 0 (No Overlap): The default assumption where clients have completely disjoint data partitions.
0.3 (Moderate Overlap): 30\% of the samples in clusters from the same data distribution are shared among clients.
0.5 (High Overlap): 50\% of the samples in clusters from the same data distribution are shared.

For consistency, we use the EMNIST dataset as the reference and fix the dirtiness level at 0.3 to isolate the impact of overlap on clustering performance.

\begin{figure*}[htbp]
    \centering
    \subfloat[][Number of Communities Found]{
    \includegraphics[width=.30\textwidth]{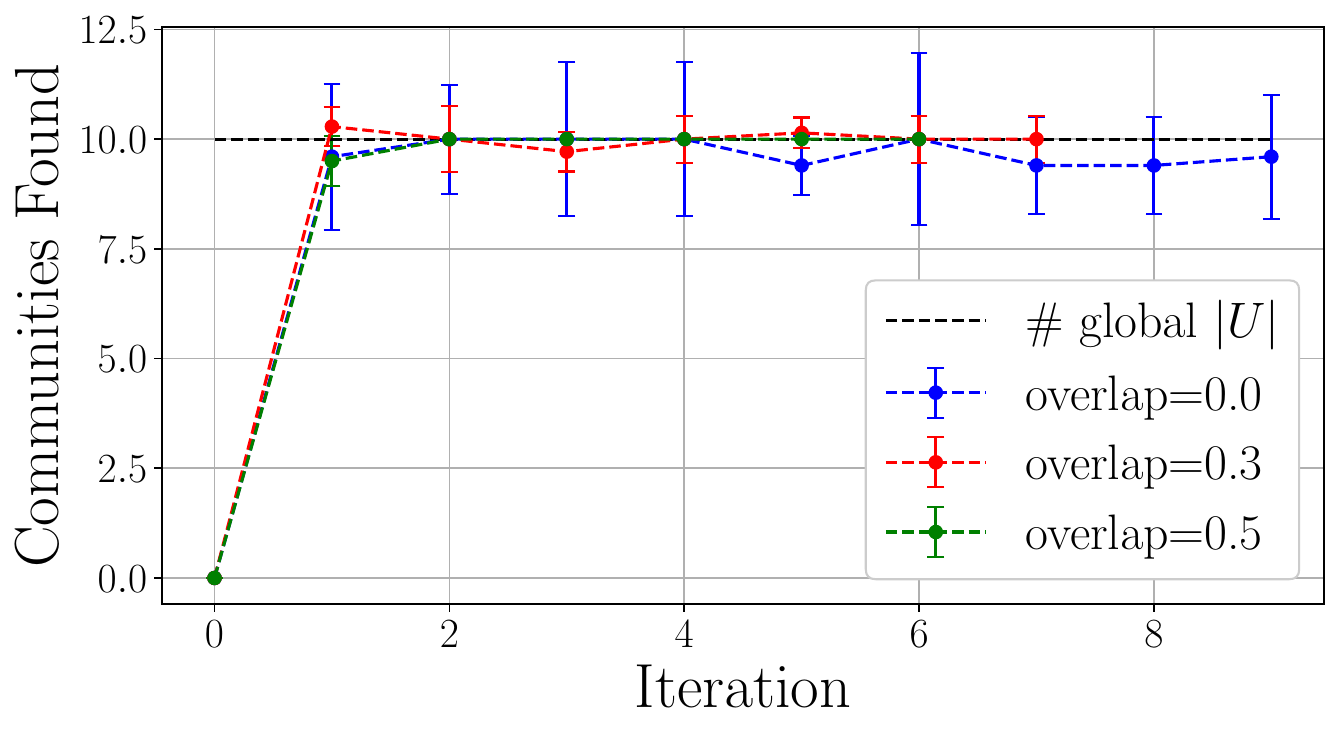}
    \label{fig:overlap:communities_found}
    }
    \subfloat[][Accuracy]{
    \includegraphics[width=.30\textwidth]{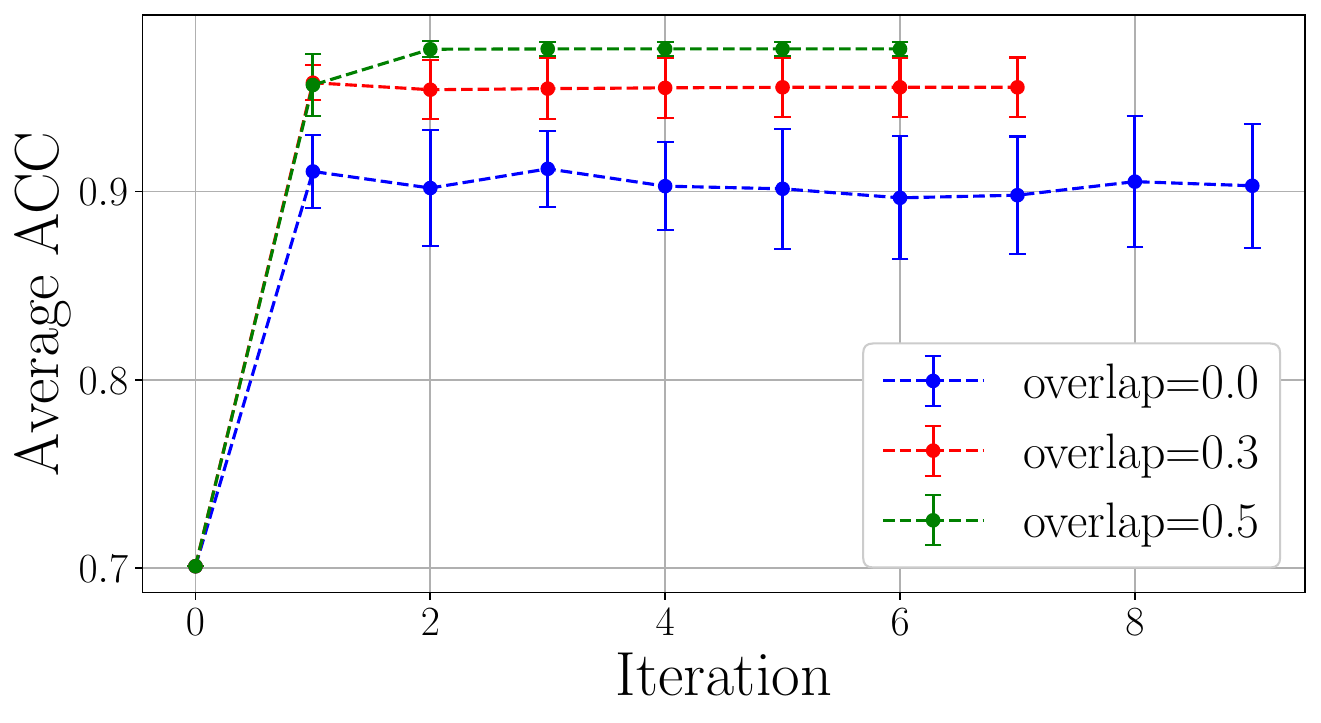}
    \label{fig:overlap:unsupervised_accuracy}
    }
    \subfloat[][Isolated clusters]{
    \includegraphics[width=.30\textwidth]{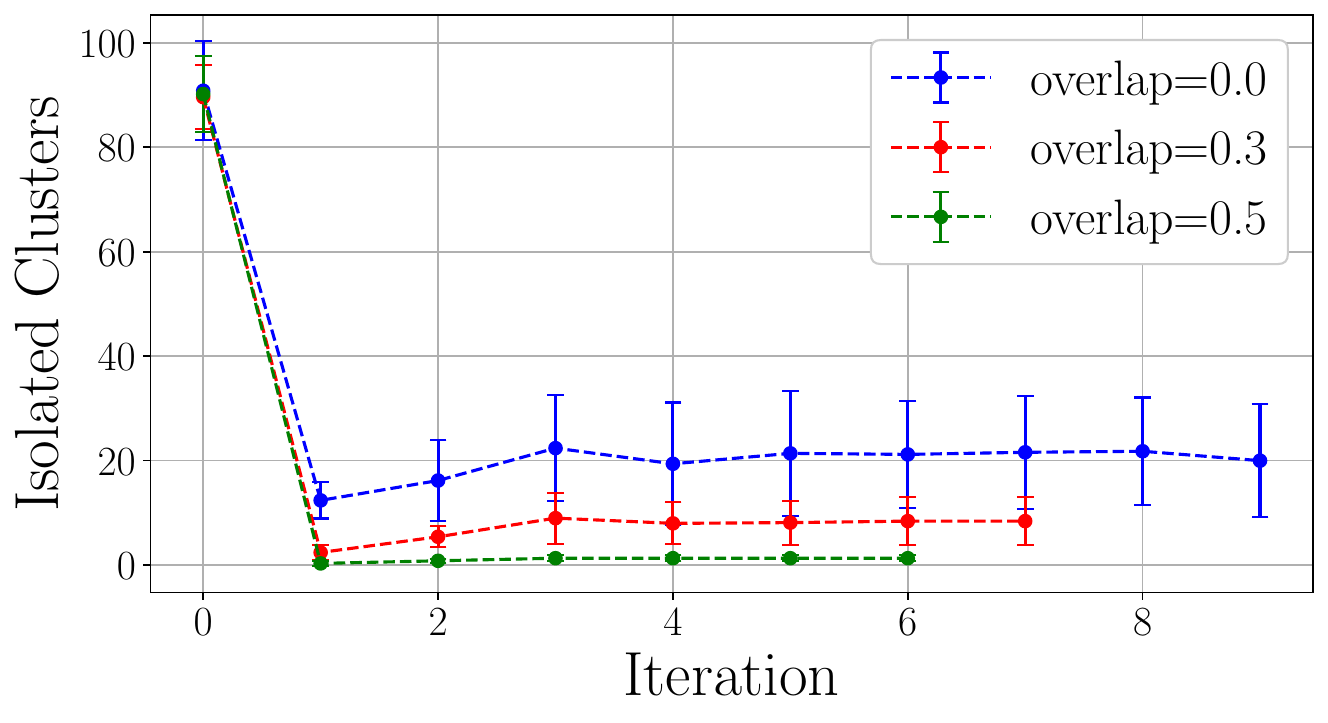}
    \label{fig:overlap:isol}
    }
    \caption{Trends in the algorithm’s performance metrics over iterations for different overlap levels: 0 (No Overlap, blue), 0.3 (Moderate Overlap, red), and 0.5 (High Overlap, green). Experiments are conducted on the EMNIST dataset using \textit{simulated clustering} with a fixed dirtiness level of 0.3 as the initial configuration. Results represent the mean ± 95\% confidence interval (CI) over five runs.} 
    \label{fig:overlap:trends}
\end{figure*}

\noindent
\textit{Results Analysis}
Across all overlap conditions, FedCRef consistently identifies the correct number of communities ($|G|$), aligned with the true number of global data distributions ($|U|$), without any incorrect associations.
The number of isolated clusters significantly decreases throughout the process, with the most substantial reduction occurring under high overlap, where isolated clusters dropped from 90.167 to 1.333. Table~\ref{tab:overlap:clusters} summarizes these clustering outcomes, while Figure~\ref{fig:overlap:trends} illustrates the trends over iterations.
Table~\ref{tab:overlap:uacc} shows that final ACC improves as data overlap increases. With no overlap, the final ACC is 0.903, rising to 0.955 with moderate overlap (0.3) and 0.976 with high overlap (0.5).
The relative accuracy improvement from the initial condition follows the same trend: $+28.82\%$ for no overlap, $+36.23\%$ for moderate overlap, and $+39.23\%$ for high overlap.
Moreover, the 95\% confidence intervals for ACC become narrower with increasing overlap, indicating greater precision in final clustering accuracy. Notably, these results closely approximate the perfect-association baseline, further validating the effectiveness of the approach.

These findings suggest that higher data overlap significantly improves both clustering accuracy and result stability in federated unsupervised learning. The observed trend is expected, as clients with shared data train models that generalize better to each other’s datasets, leading to more precise associations and improved local clustering.
However, it is important to highlight that FedCRef achieves strong performance even in the most challenging scenario (0 overlap), where clients have completely disjoint data partitions. This demonstrates the method’s resilience in fully decentralized settings, where no data is shared between clients.

\begin{table}[htbp]
\centering
\caption{Algorithm outcomes across different overlap levels (mean values over five runs). Experiments were conducted on the EMNIST dataset with client data partitioning initialized at a dirtiness level of 0.3.}
\label{tab:overlap:clusters}
\begin{adjustbox}{max width=\columnwidth}   
\begin{tabular}{cccccc}
\toprule
 ACC init & Overlap  & Comm. Found / & Wrong Ass. & Isolated \\
 & &  True Categories $|U|$ & & Start $\rightarrow$ End \\
\midrule
 \multirow{3}{*}{0.7}   
         & 0   &  $9.8 (\pm 1.7)$ / 10 & 0.0 & $88.2(\pm 3.2)  \rightarrow 18.8 (\pm 9.3)$ \\
        & 0.3 &  $10 (\pm 0.4)$ / 10 & 0.0 &  $89.571(\pm 5.8) \rightarrow 8.429 (\pm 3.2)$ \\
        & 0.5 &  $10 (\pm 0.1)$ / 10 & 0.0 & $90.167 (\pm 4.4) \rightarrow 1.333 (\pm 0.8)$ \\
\bottomrule
\end{tabular} \end{adjustbox}

\end{table}

\begin{table}[htbp]
\centering
\caption{Final ACC and ACC improvement (mean ± 95\% CI) across different overlap levels. Experiments were conducted on the EMNIST dataset with client data partitioning initialized at a dirtiness level of 0.3.}
\label{tab:overlap:uacc}
\begin{adjustbox}{max width=\columnwidth}   
\begin{tabular}{ccccc}
\toprule
 ACC Init & Overlap & ACC End  & Improvement (\%) & ACC End \\
         & & (Our) & & (Perfect Associations) \\
\midrule
 \multirow{3}{*}{0.7}   
        & 0   & $0.903  (\pm 0.033)$ & $28.816  (\pm 3.16)$ & $0.953 (\pm 0.015) $\\
             & 0.3 & $0.955  (\pm 0.016)$ & $36.234  (\pm 2.32)$ & $0.962 (\pm 0.012)$\\
             & 0.5 & $0.976 (\pm 0.004)$ & $39.23  (\pm 1.12)$ & $0.97 (\pm 0.003) $\\
\bottomrule
\end{tabular} \end{adjustbox}
%
\end{table}

\subsection{Impact of the number of clients }
\label{sec:exp:numclients}

\textit{Experimental Setup}.
In this final set of experiments, we assess the scalability of \textit{FedCRef} by evaluating its performance under increasing system sizes. Specifically, we vary the number of participating clients (N) to observe how the algorithm copes with the growing complexity of a larger federated network. As in previous experiments, we use the EMNIST dataset and fix the dirtiness level at 0.3, corresponding to an average initial clustering accuracy of approximately 0.7 per client.

Note that since each client is initialized with the same number of samples (as defined by the experiment’s default parameters), the total number of global samples increases linearly with the number of clients.

\begin{figure*}[htbp]
    \centering
    \subfloat[][Number of Communities Found]{
    \includegraphics[width=.24\textwidth]{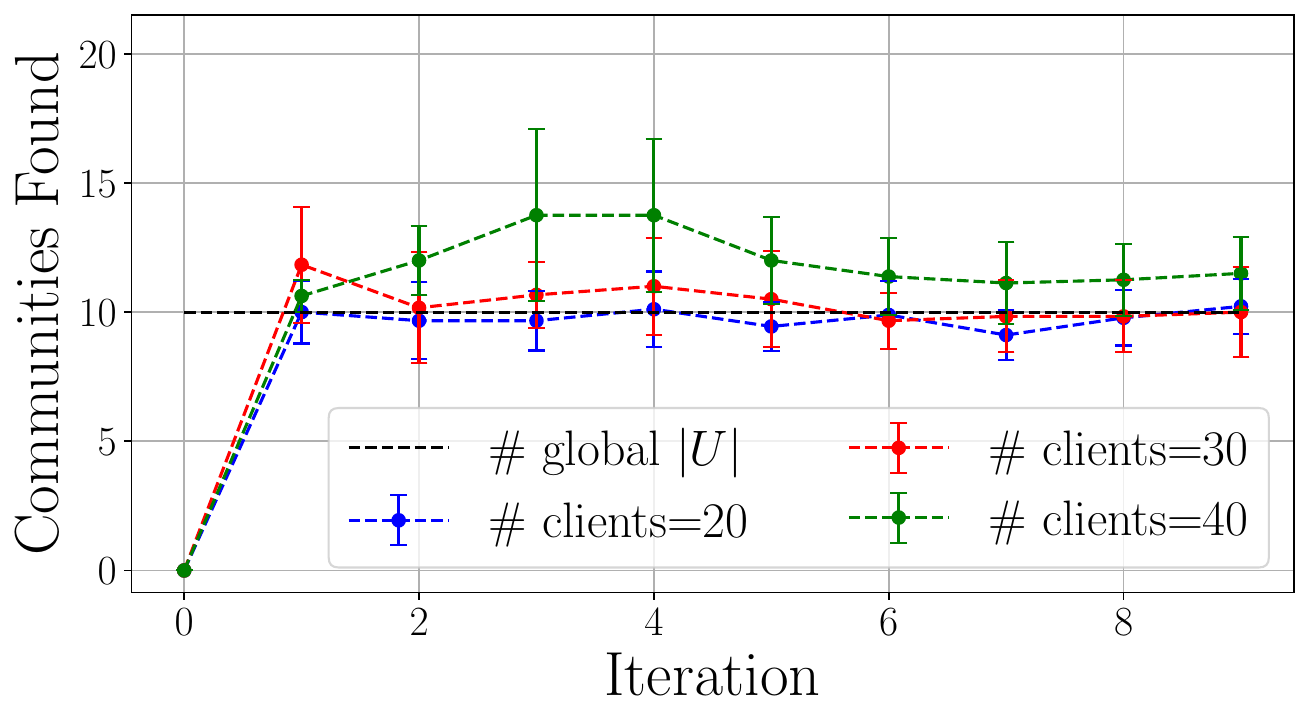}
    \label{fig:numclients:communities_found}
    }
    \subfloat[][Accuracy]{
    \includegraphics[width=.24\textwidth]{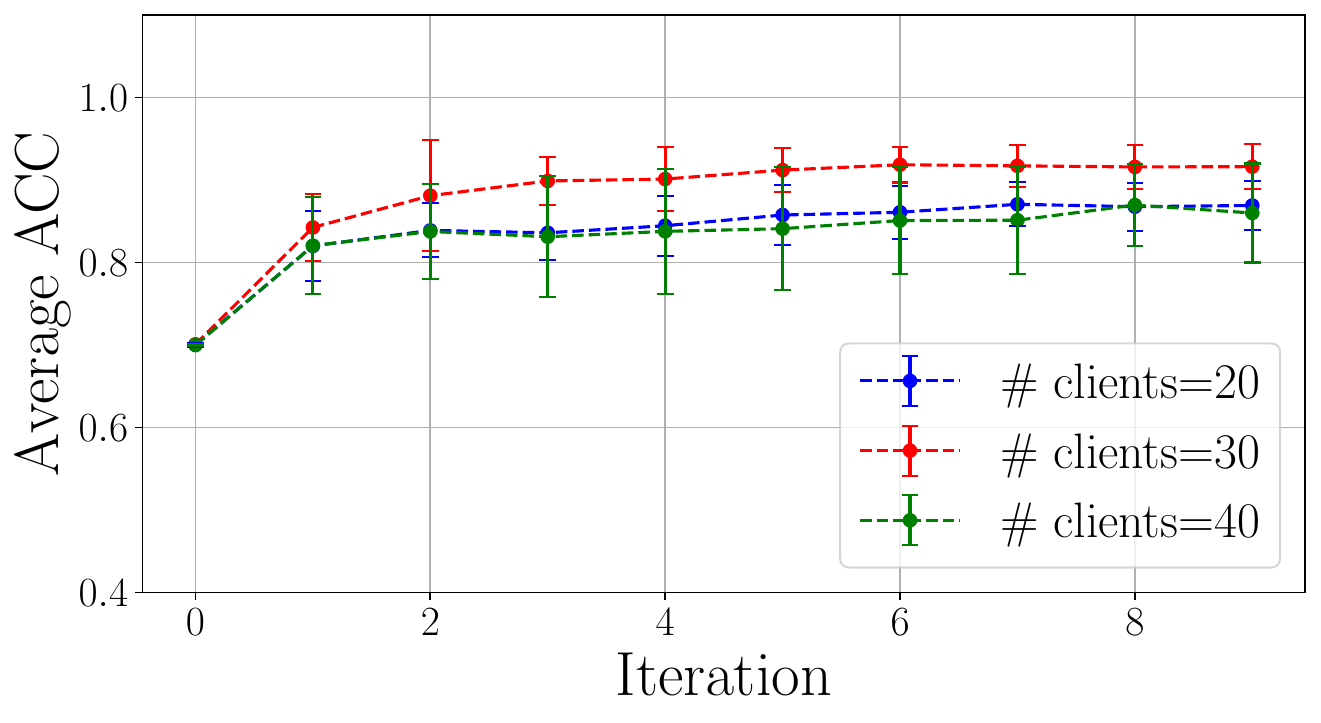}
    \label{fig:numclients:unsupervised_accuracy}
    }
    \subfloat[][Wrong associations (\%)]{
    \includegraphics[width=.24\textwidth]{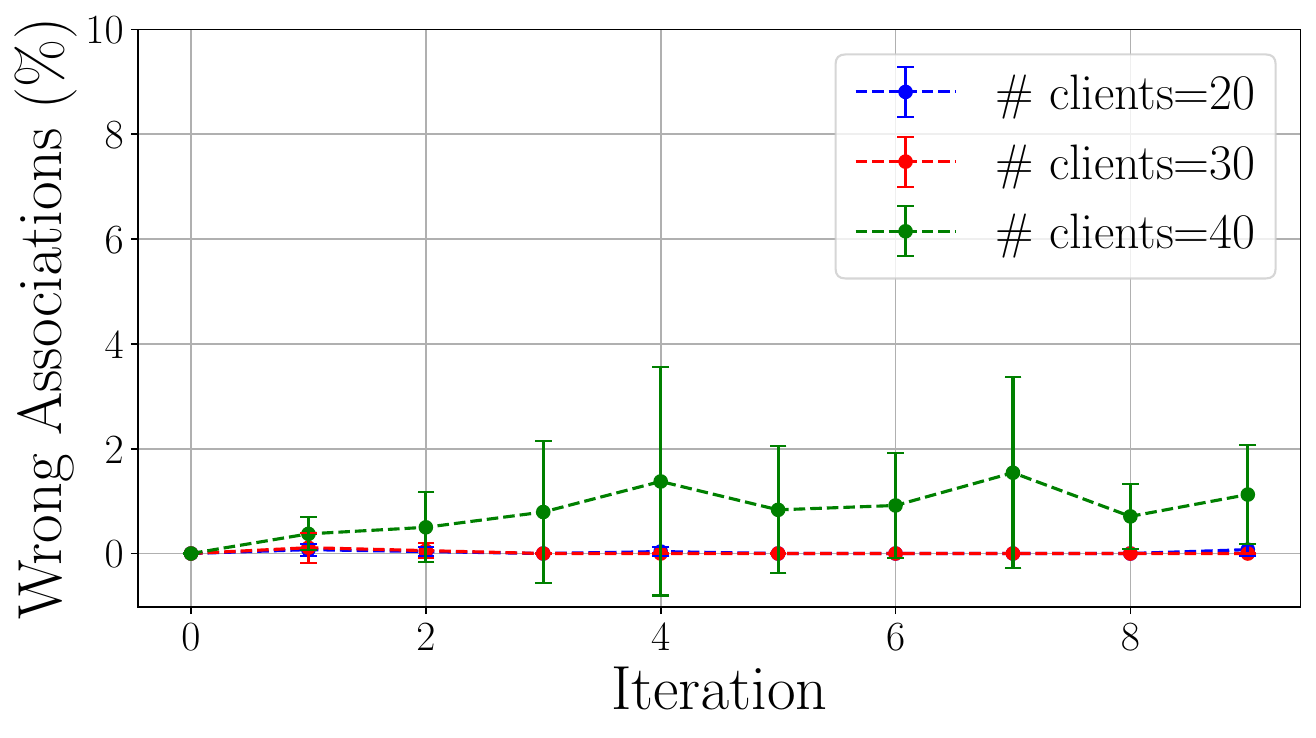}
    \label{fig:numclients:wrong}
    }
    \subfloat[][Isolated clusters]{
    \includegraphics[width=.24\textwidth]{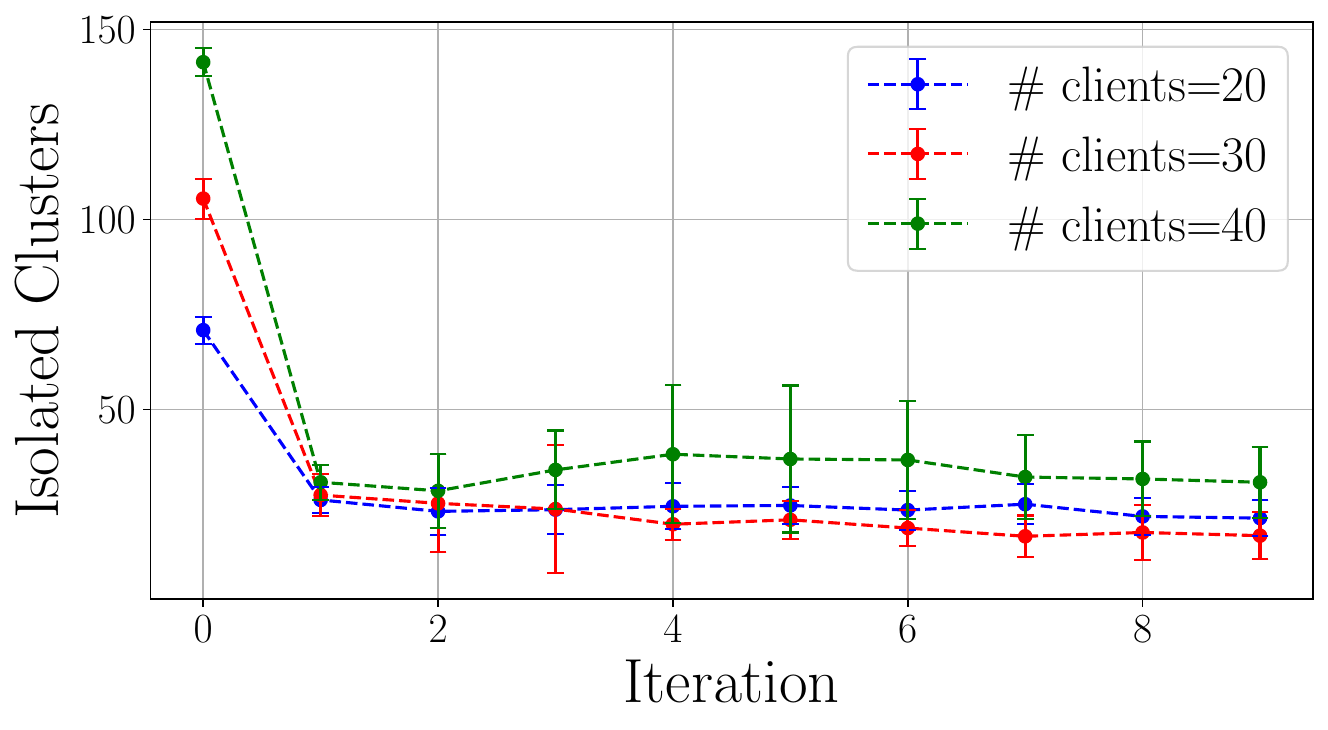}
    \label{fig:numclients:isol}
    }
    \caption{Trends in the algorithm’s performance metrics over iterations for different numbers of clients per simulation. Tests on the EMNIST dataset and fixed dirtiness at 0.3 as the initial configuration. Results represent the mean ± 95\% confidence interval (CI) over five runs.}
    \label{fig:numclients:trends}
\end{figure*}

\noindent
\textit{Results Analysis.} 
FedCRef scales effectively up to 40 clients, maintaining strong alignment between the discovered communities and the true data distributions. As shown in Table~\ref{tab:numbers:clusters}, the system achieves peak performance at 30 clients, likely because the heterogeneity of the dataset has already been well captured at that scale, allowing the algorithm to form highly accurate associations.
At 40 clients, while the performance remains strong, we observe a slight increase in the number of data distributions detected—marginally exceeding the true count—as well as a small number of incorrect associations. This can be attributed to the increased complexity of the clustering space: with 40 clients, the total number of local data splits (i.e., starting isolated clusters) rises to approximately 140, requiring the algorithm to establish a larger number of associations across clients. Despite this, overall accuracy remains high, and the method continues to identify nearly all true data categories correctly. Figure~\ref{fig:numclients:trends} illustrates detailed trends over iterations.

We also observe that, when perfect associations are assumed, increasing the number of clients (i.e., the overall global data) leads to higher final accuracy, confirming that the federated training process is effective, as show in the last column of Table~\ref{tab:numbers:uacc}. The main challenge lies in reliably discovering high-quality associations between client clusters.

These results suggest that FedCRef handles increasing client populations well, with minor trade-offs in precision at higher scales.

\begin{table}[htbp]
\centering
\caption{Mean Clustering Outcomes at varying the number of clients of the system. Tests on the EMNIST dataset and fixed dirtiness at 0.3 as the initial configuration.}
\label{tab:numbers:clusters}

\begin{adjustbox}{max width=\linewidth}   \begin{tabular}{cccccc}
\toprule
 ACC init & Clients (\#) & Comm. Found / & Wrong Ass. & Isolated \\
 & & True Categories $|U|$ & (\%) & Start $\rightarrow$ End \\
\midrule
 \multirow{3}{*}{0.7}   
& 20  & $10.2(\pm 1.1) $ / 10 & $0.222 (\pm 0.339) $ & $ 70.9(\pm 3.5)  \rightarrow  21.4 (\pm 4.7) $ \\
 & 30  & $10.0(\pm 1.76) $ / 10 & $0.0 (\pm 0.0) $ & $105.5(\pm 5.3)  \rightarrow  16.8 (\pm 6.2) $ \\
 & 40  & $11.5(\pm 1.41) $ / 10 & $ 1.12 (\pm 0.941)$ & $141.4(\pm 3.7)  \rightarrow  30.9 (\pm 9.4) $ \\
\bottomrule
\end{tabular} \end{adjustbox}
\end{table}

\begin{table}[htbp]
\centering
\caption{Mean ± 95\% CI  final Accuracy (ACC) and Improvement varying the number of clients of the system. Tests on the EMNIST dataset and fixed dirtiness at 0.3 as the initial configuration.}
\label{tab:numbers:uacc}
\begin{adjustbox}{max width=\linewidth}   \begin{tabular}{cccccc}
\toprule
 ACC Init & Clients (\#) & ACC End & Improvement (\%) & Acc End\\
   &  & (Our) & & (Perfect Associations)\\
\midrule
 \multirow{3}{*}{0.7}       
& 20  & $ 0.869 (\pm 0.03) $ & $ 24.143 (\pm 4.286) $ & $ 0.921 (\pm 0.041) $ \\
& 30 & $ 0.916  (\pm 0.027)$ & $ 30.67 (\pm 3.88)$  & $ 0.937 (\pm 0.029) $ \\
& 40  & $ 0.86 (\pm 0.06) $ & $ 22.857 (\pm 8.571) $ & $ 0.953 (\pm 0.007) $ \\
\bottomrule
\end{tabular} \end{adjustbox}

\end{table}

We further use this set of experiments to assess the number of iterations required for convergence and the corresponding runtime as the number of clients increases.
Although the stopping criterion of FedCRef is defined heuristically, the algorithm consistently converged in all experimental runs and configurations.
In all cases, most clients stabilized within 5–8 iterations (a typical trend of active clients is shown in Fig.\ref{fig:emnist:active_clients}), while the global stopping criterion (i.e., \textit{small} changes in isolated clusters or formed groups) ensures convergence for the remaining few clients.
Table\ref{tab:convergence} reports average values over multiple runs, showing that FedCRef converges within a small number of iterations across all settings (up to 20 in the largest scenario). Note that in Fig.~\ref{fig:numclients:trends}, the plots are truncated once the trends become stable, to ease comparison across different client settings.
While the theoretical complexity of the association step (i.e., pairwise inferences) is quadratic in the number of clients, in our experiments (N = 20–40) the observed runtime increases linearly.
We attribute this to the parallel execution of local training, early stabilization of most clients, and the relatively small scale of the testbed.

\begin{table}
\centering
\caption{Convergence statistics of FedCRef for different numbers of clients.}
\label{tab:convergence}
\begin{adjustbox}{max width=\linewidth} 
\begin{tabular}{ccc}
\toprule
\# Clients (N) & Iterations to Converge & Runtime (min) \\
\midrule
20 & $11.5 \pm 2.1 $ & $50.4 \pm 11.8$ \\
30 & $13.7 \pm 1.7 $ & $71.2 \pm 9.9$ \\
40 & $16.7 \pm 3.6 $ & $87.6 \pm 16.2$ \\
\bottomrule
\end{tabular}
\end{adjustbox}
\end{table}


%% file: 06_conclusion.tex
\section{Conclusion}
\label{sec:conclusion}
In this study, we propose FedCRef, a novel unsupervised federated learning framework for federated clustering. It operates under realistic settings with heterogeneous client data from multiple unknown distributions. Unlike prior work, FedCRef does not require knowing the number of global distributions (K) in advance, i.e., it discovers them while learning specialized representations for each.
FedCRef integrates: (i) federated cluster-wise training, where clients jointly train models on clusters sharing the same distribution, and (ii) iterative local clustering refinement, which improves cluster quality and global alignment. Without centralized coordination, clients converge to a consistent clustering structure while keeping data local.
We evaluate FedCRef on four public datasets borrowed from computer vision and compared the results to real centralized (DEC) and federated (UIFCA) clustering benchmarks, as well as upper-bound baselines using perfect associations and clean initializations. FedCRef consistently identified the correct number of clusters and achieved up to 0.95 ACC within a few iterations.
We further analyze scalability and sensitivity to data-overlap, finding that moderate overlap improves convergence and robustness—even with noisy initialization. Finally, we that our method is easily configurable and moderately sensitive to hyper-parameter tuning. Future work will explore adaptive associations, hierarchical clustering, and cross-client representation sharing to scale FedCRef to larger, more heterogeneous systems.